\documentclass{article} 
\usepackage{iclr2022_conference,times}

\usepackage{amsmath,amsfonts,bm}

\def\eqref#1{equation~\ref{#1}}

\def\1{\bm{1}}

\DeclareMathAlphabet{\mathsfit}{\encodingdefault}{\sfdefault}{m}{sl}
\SetMathAlphabet{\mathsfit}{bold}{\encodingdefault}{\sfdefault}{bx}{n}

\usepackage{hyperref}
\usepackage{url}

\usepackage{lineno}
\usepackage{latexsym}
\usepackage{float}
\usepackage{microtype}

\usepackage{amsmath}
\usepackage{amssymb}
\usepackage{bm}
\usepackage{multirow}
\usepackage{makecell, diagbox}
\usepackage{subfigure}
\usepackage{booktabs}
\usepackage{collcell}
\usepackage{etoolbox}
\usepackage{pgf}
\usepackage{xspace}
\newcommand\blfootnote[1]{%
  \begingroup
  \renewcommand\thefootnote{}\footnote{#1}%
  \addtocounter{footnote}{-1}%
  \endgroup
}

\newcommand{\method}{\textsc{X-Mixup}\xspace}

\title{Enhancing Cross-lingual Transfer by Manifold Mixup}

\iclrfinalcopy

\author{Huiyun Yang$^{1}$, Huadong Chen$^{1}$, Hao Zhou\thanks{Corresponding author.}$^{~~1}$, Lei Li$^{2}$ \\
$^{1}$ByteDance AI Lab \\
$^{2}$University of California, Santa Barbara \\
\texttt{\{yanghuiyun.11,chenhuadong.howard,
zhouhao.nlp\}@bytedance.com} \\ \texttt{leili@cs.ucsb.edu}
}

\begin{document}

\maketitle

\begin{abstract}
Based on large-scale pre-trained multilingual representations, recent cross-lingual transfer methods have achieved impressive transfer performances.
However, the performance of target languages still lags far behind the source language.
In this paper, our analyses indicate such a performance gap is strongly associated with the cross-lingual representation discrepancy.
To achieve better cross-lingual transfer performance, we propose the cross-lingual manifold mixup (\method) method, which adaptively calibrates the representation discrepancy and gives compromised representations for target languages.
Experiments on the XTREME benchmark show \method achieves 1.8\% performance gains on multiple text understanding tasks, compared with strong baselines, and reduces the cross-lingual representation discrepancy significantly.\blfootnote{Code is available at \url{https://github.com/yhy1117/X-Mixup}.}

\end{abstract}

\section{Introduction}
Many natural language processing tasks have shown exciting progress utilizing deep neural models. 
However, these deep models often heavily rely on sufficient annotation data, which is not the case in the multilingual setting.
The fact is that most of the annotation data are collected for popular languages like English and Spanish~\citep{Ponti2019,Joshi2020}, while many long-tail languages could hardly obtain enough annotations for supervised training.
As a result, cross-lingual transfer learning \citep{PrettenhoferS2011,Wan2011,Ruder2019} is crucial, transferring knowledge from the annotation-rich \textit{source language} to low-resource or zero-resource \textit{target languages}. In this paper, we focus on the zero-resource setting, where labeled data are only available in the source language.

Recently, multilingual pre-trained models \citep{xlm,xlmr,mt5} offer an effective way for cross-lingual transfer, which yield a universal embedding space across various languages.
Such universal representations make it possible to transfer knowledge from the source language to target languages through the embedding space, significantly improving the transfer learning performance~\citep{Chen2019,Zhou2019,keung2019,Filter}.
Besides, \citet{xnli} proposes \textit{translate-train}, a simple yet effective cross-lingual data augmentation method, which constructs pseudo-training data for each target language via machine translation.
Although these works have achieved impressive improvements in cross-lingual transfer~\citep{xtreme,xtreme-r}, significant performance gaps between the source language and target languages still remain (see Table \ref{table1}).
\citet{xtreme} refers to the gap as the \textit{cross-lingual transfer gap}, a difference between the performance on the source and target languages.

\begin{figure}[t]
\centering
\subfigure[Translate-train]{
\label{fig_intro_a}
\includegraphics[width=0.3\textwidth]{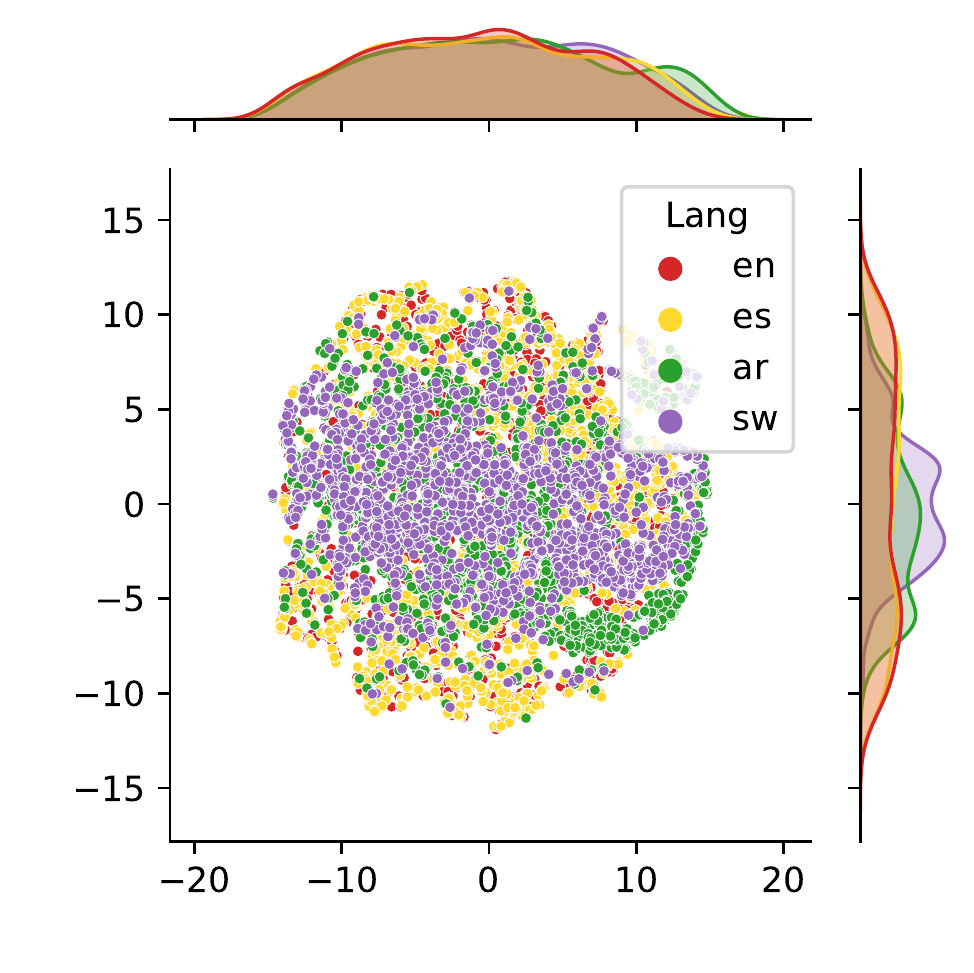}}
\qquad
\subfigure[\method]{
\label{fig_intro_b} 
\includegraphics[width=0.3\textwidth]{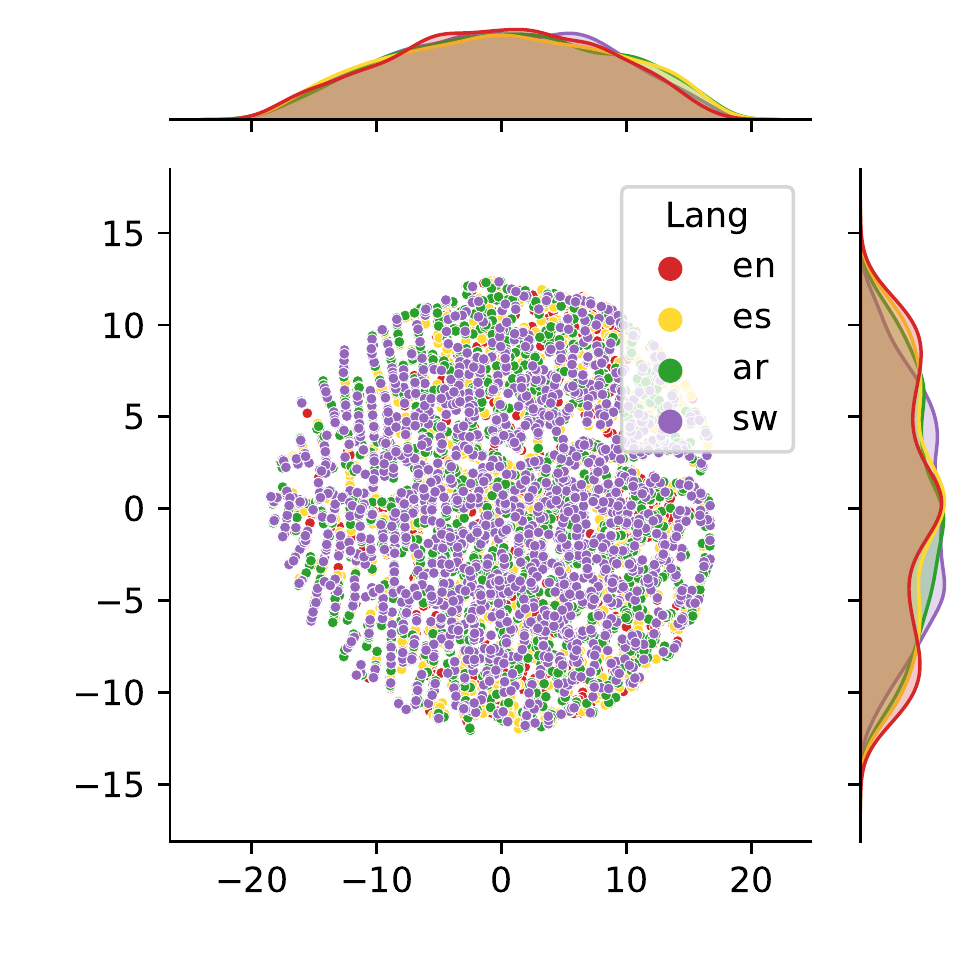}}
\caption{Representation visualization of four languages: English (en), Spanish (es), Arabic (ar) and Swahili (sw) based on XLM-R.
We plot the sentence representation of the XNLI test set, which is parallel across 15 languages. We average hidden states of the last layer to get sentence representations and implement the dimensionality reduction by PCA.
Obviously, the cross-lingual representation discrepancies are large in translate-train, but \method reduces the discrepancy significantly.}
\label{fig_intro}
\end{figure}
To investigate how the cross-lingual transfer gap emerges,
we perform relevant analyses, demonstrating that transfer performance correlates well with the \textit{cross-lingual representation discrepancy}~(see Section \ref{motivation} for details).
Here the cross-lingual representation discrepancy means the degree of difference between the source and target language representations in the universal embedding space.
As shown in Figure \ref{fig_intro_a}, in translate-train, the representation distribution of Spanish almost overlaps with English, while Arabic shows a certain representation discrepancy compared with English and Swahili performs larger discrepancy, where translate-train achieves 88.6 accuracy on English, 85.7 on Spanish, 82.2 on Arabic and 77.0 on Swahili.
Intuitively, a larger representation discrepancy could lead to a worse cross-lingual transfer performance.

In this paper, we propose the \textit{Cross-Lingual Manifold Mixup}~(\method) approach to fill the cross-lingual transfer gap.
Based on our analyses, reducing the cross-lingual representation discrepancy is a promising way to narrow the transfer gap.
Given the cross-lingual representation discrepancy is hard to remove, \method directly faces the issue and explicitly accommodates the representation discrepancy in the neural networks, by mixing the representation of the source and target languages during training and inference.
With \method, the model itself can learn how to escape the discrepancy, which adaptively calibrates the representation discrepancy and gives compromised representations for target languages to achieve better cross-lingual transfer performance.
\method is motivated by robust deep learning \citep{vincent2008extracting}, while \method adopts the mixup \citep{Zhang2018} idea to handle the cross-lingual discrepancy.

Specifically, \method is designed upon the translate-train approach, faced with the exposure bias \citep{Ranzato2016} problem and data noise problem.
During training, the source sequence is a real sentence and the target sequence is a translated one, while situations are opposite during inference.
Besides, the translated text often introduces some noises due to imperfect machine translation systems.
To address them, we further impose the \textit{Scheduled Sampling}~\citep{Bengio2015} and \textit{Mixup Ratio} in \method to handle the distribution shift problem and data noise problem, respectively.

We verify \method on the cross-lingual understanding benchmark XTREME \citep{xtreme}, which includes several understanding tasks
and covers 40 languages from diverse language families.
Experimental results show \method achieves 1.8\% performance gains across different tasks and languages, comparing with strong baselines.
It also reduces the cross-lingual representation discrepancy significantly, as Figure \ref{fig_intro_b} shows.

\section{Related Work}
\textbf{Multilingual Representation Learning}\quad
Recent studies have demonstrated the superiority of large-scale pre-trained multilingual representations on downstream tasks.

Multilingual BERT (mBERT; \citealp{bert}) is the first work to extend the monolingual pre-training to the multilingual setting.
Then, several extensions achieve better cross-lingual performances by introducing more monolingual or parallel data and new pre-training tasks, such as Unicoder \citep{unicoder}, XLM-R \citep{xlmr}, ALM \citep{Yang2020}, MMTE \citep{MMTE}, InfoXLM \citep{infoxlm}, HICTL \citep{hictl}, ERNIE-M \citep{erniem}, mT5 \citep{mt5}, nmT5 \citep{nmt5}, AMBER \citep{amber} and VECO \citep{veco}. They have been the standard backbones of current cross-lingual transfer methods.

\textbf{Cross-lingual Transfer Learning}\quad
Cross-lingual transfer learning \citep{PrettenhoferS2011,Wan2011,Ruder2019} aims to transfer knowledge learned from source languages to target languages.
According to the type of transfer learning \citep{Pan2010}, previous cross-lingual transfer methods can be divided into three categories: instance transfer, parameter transfer, and feature transfer.
The cross-lingual transferability improves a lot when engaged with the instance transfer by translation (i.e. translate-train, translate-test) or other cross-lingual data augmentation methods~\citep{Jasdeep2019,Mihaela2020,Qin2020,zheng2021}.
\citet{Chen2019} and \citet{Zhou2019} focus on the parameter transfer to learn a share-private model architecture. Besides, other works implement the feature transfer to learn the language-invariant features by adversarial networks \citep{keung2019,Chen2019} or re-alignment \citep{libovicky2020,zhao2020}.
\method utilizes both the instance transfer and feature transfer, which is based on the translate-train data augmentation approach and implements the feature transfer by cross-lingual manifold mixup.

\textbf{Mixup and Its Variants}\quad
Mixup \citep{Zhang2018} proposes to train models on the linear interpolation at both the input level and label level, which is effective to improve the model robustness and generalization.
Generally, the interpolated pair is selected randomly.
Manifold mixup \citep{Verma2019} performs the interpolation in the latent space by conducting the linear combinations of hidden states.
Previous mixup methods \citep{chen-etal-2020-mixtext,Jindal2020} focus on the monolingual setting. However, X-Mixup focuses on the cross-lingual setting and faces many new challenges (see Section \ref{method} for details).
Besides, in contrast to previous mixup methods, \method mixes the parallel pairs, which share the same semantics across different languages.
As a result, the choice of parallel pairs for interpolation can build a smart connection between the source and target languages.

\section{Analyses of the Cross-lingual Transfer Performance}
\label{motivation}
In this section\footnote{In our analyses, we take English as the source language, and the dissimilar language is the language which is dissimilar to English.}, we concentrate on the cross-lingual transfer performance and find it is strongly associated with the cross-lingual representation discrepancy.
Firstly, we observe the cross-lingual transfer performance on different target languages and propose an assumption.
Then we conduct qualitative and quantitative analyses to verify it. 

\begin{table}[t]
\setlength{\belowcaptionskip}{0.2cm}
\caption{Cross-lingual transfer performances of POS and NER tasks on languages with different data resources or different language families, where there are only labeled training data in English. 
The data resource refers to the resource of each language utilized in the pre-training process.
For the language family, English belongs to the Germanic languages, so we divide languages into two types: Germanic one and others.
Results show high-resource languages outperform low-resource ones significantly and languages dissimilar to the source language tend to perform worse.}
\centering
\resizebox{\linewidth}{!}{
\begin{tabular}{c|c|c||c c c|c c c||c c c|c c c}
\toprule
\multicolumn{2}{c|}{\multirow{2}{*}{Language Type}} & \multirow{2}{*}{Source} & \multicolumn{6}{c||}{Language Resource} & \multicolumn{6}{c}{Language Family} \\
\cline{4-15}
\multicolumn{2}{c|}{} & &  \multicolumn{3}{c|}{High-resource} & \multicolumn{3}{c||}{Low-resource} &
\multicolumn{3}{c|}{Germanic} & \multicolumn{3}{c}{Other} \\
\hline
\multicolumn{2}{c|}{Language} & en & es & it & pt & eu & kk & mr & af & de & nl & ar & hi & ja \\
\hline
\multirow{2}{*}{mBERT} & POS & 95.5 & 86.9 & 88.4 & 86.2 & 60.7 & 70.5 & 69.4 & 86.6 & 85.2 & 88.6 & 56.2 & 67.2 & 49.2 \\
 & NER & 85.2 & 77.4 & 81.5 & 80.8 & 66.3 & 45.8 & 58.2 & 77.4 & 78.0 & 81.8 & 41.1 & 65.0 & 29.0\\
\hline
\multirow{2}{*}{XLM-R} & POS & 96.1 & 88.3 & 89.4 & 87.6 & 72.5 & 78.1 & 80.8 & 89.8 & 88.5 & 89.5 & 67.5 & 76.4 & 15.9 \\
 & NER & 84.7 & 79.6 & 81.3 & 81.9 & 60.9 & 56.2 & 68.1 & 78.9 & 78.8 & 84.0 & 53.0 & 73.0 & 23.2\\
\bottomrule
\end{tabular}
}
\label{table1}
\end{table}

Although previous studies \citep{xtreme,xtreme-r} have shown impressive improvements on cross-lingual transfer, the cross-lingual transfer gap is still pretty large, more than 16 points in \citet{xtreme}.
Furthermore, results in Table \ref{table1} show \textbf{the performance of low-resource languages and dissimilar languages fall far behind other languages in cross-lingual transfer tasks.}

\begin{table}[t]
\setlength{\belowcaptionskip}{0.2cm}
\caption{Spearman’s rank correlation $\rho$ between the CKA score and cross-lingual transfer performance on two XTREME tasks, where $^{\dagger}$ denotes training on the source language, and $^{\ddagger}$ denotes the translate-train approach. $^*$ denotes the p-value is lower than 0.05. Results indicate the correlation is solid.
}
\centering
\begin{tabular}{c c c c c}
\toprule
Task & XNLI$^{\dagger}$ & XNLI$^{\ddagger}$ & PAWS-X$^{\dagger}$ & PAWS-X$^{\ddagger}$ \\
\midrule
$\rho$ & $0.76^*$ & $0.69^*$ & $0.90^*$ & $0.93^*$ \\
\bottomrule
\end{tabular}
\label{table_spear}
\end{table}
Compared with English, the representations of other languages, especially low-resource languages, are not well-trained \citep{Lauscher2020,Wu2020}, because high-resource languages dominate the representation learning process, which results in the cross-lingual representation discrepancy.
Besides, dissimilar languages often show differences in language characteristics (like vocabulary, word order), which also leads to the representation discrepancy.
As a result, we assume that the cross-lingual transfer performance is closely related to the representation discrepancy between the source language and target languages.

Following \citet{conneau2020}, we utilize the linear centered kernel alignment (CKA; \citealp{Kornblith2019}) score to indicate the cross-lingual representation discrepancy
\begin{equation}
\text{CKA}(X,Y)=\frac{||Y^\top X||^2_\text{F}}{||X^\top X||^2_\text{F}||Y^\top Y||^2_\text{F}},
\end{equation}
where X and Y are parallel sequences from the source and target languages, respectively.
A higher CKA score denotes a smaller cross-lingual representation discrepancy.

To verify our assumption, we perform qualitative and quantitative analyses on the relationship between the CKA score and cross-lingual transfer performance.
Figure \ref{fig_cka_line} in Appendix \ref{appendix_analysis} indicates a higher CKA score tends to induce better cross-lingual transfer performance.
We also calculate the Spearman’s rank correlation between the CKA score and the transfer performance in Table \ref{table_spear}, which shows a strong correlation between them.
Both the trend and correlation score confirm \textbf{the cross-lingual transfer performance is highly related to the cross-lingual representation discrepancy}.

\section{Methodology: \method}
\label{method}
Based on the aforementioned analyses, we believe that reducing the cross-lingual representation discrepancy is the key to filling the cross-lingual transfer gap. 
In this section, we propose \method to explicitly reduce the representation discrepancy by implementing the manifold mixup between the source language and target language.
With \method, the model can adaptively calibrate the representation discrepancy and give compromised representations for target languages.
This section will first introduce the overall architecture of \method and its details. After that, the training objectives and inference process will be shown.

\subsection{Overall Architecture}
\begin{figure}[t]
\centering
\includegraphics[scale=0.65]{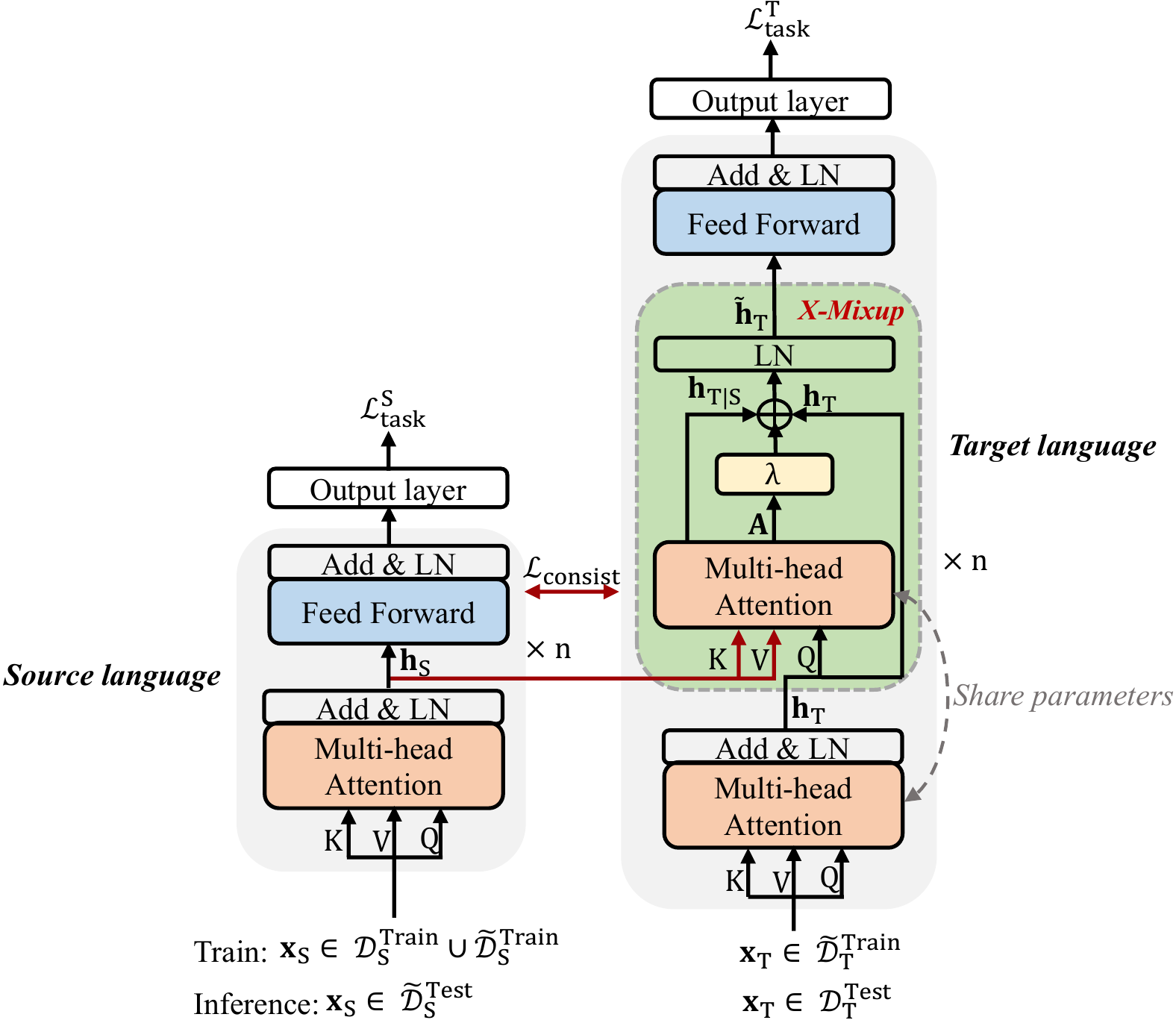}
\caption{The model architecture of \method, where the cross-lingual manifold mixup process is in the green block. Note that the manifold mixup process is implemented only in a certain layer (the same layer of both sides), and in other layers the process is omitted. 
}
\label{fig_xmixup}
\end{figure}

Figure \ref{fig_xmixup} illustrates the overall architecture of X-Mixup.
Sequences from the source and target languages are first encoded separately.
Then within the encoder, \method implements the manifold mixup between the paired sequences  (original sequence and its translation) within a specific layer, where \textit{Mixup Ratio} controls the degree of mixup and \textit{Scheduled Sampling} schedules the data sampling process during training.

\textbf{Notations}\quad
We use $S$ to denote the source language and $T$ to denote the target language. $\bm{h}^l$ denotes the hidden states of a sequence in layer $l$.
$\mathcal{D}$ denotes the real text data collection and $\tilde{\mathcal{D}}$ denotes the translation data collection.
For downstream understanding tasks, there are annotation data in the source language $\mathcal{D}_S^{\text{Train}}=(\mathcal{X}_S^{\text{Train}}, \mathcal{Y}_S^{\text{Train}})$ and raw test data in the target language $\mathcal{D}_T^{\text{Test}}=(\mathcal{X}_T^{\text{Test}})$. Through translate-train, we can get pseudo-training data in the target language
$\tilde{\mathcal{D}}_T^{\text{Train}}=(\tilde{\mathcal{X}}_T^{\text{Train}}, \tilde{\mathcal{Y}}_T^{\text{Train}})$.
Similarly, through translate-test, we can get pseudo-test data in the source language $\tilde{\mathcal{D}}_S^{\text{Test}}=(\tilde{\mathcal{X}}_S^{\text{Test}})$. 
During training, the Scheduled Sampling process uses translation data\footnote{These data are acquired by forward translation (from $S$ to $T$) then backward translation (from $T$ to $S$).} $\tilde{\mathcal{D}}_S^{\text{Train}}=(\tilde{\mathcal{X}}_S^{\text{Train}})$ from the source language.
Note that we use translation data ($\tilde{\mathcal{X}}_T^{\text{Train}}$ and $\tilde{\mathcal{X}}_S^{\text{Test}}$) and translate-train labels ($\tilde{\mathcal{Y}}_T^{\text{Train}}$) from the official XTREME repository, which is in the same setting as baselines.

\textbf{Basic Model}\quad
We use mBERT \citep{bert} or XLM-R \citep{xlmr} as the backbone model.
Within each layer, there are two sub-layers: the multi-head attention layer and the feed-forward layer\footnote{In this section, the feed-forward layer is omitted for simplification.}, followed by the residual connection and layer norm. We use the same multi-head attention layer (see details in Appendix \ref{multi_head}) as BERT \citep{bert}, where inputs are query, key, and value respectively.
In layer $l+1$, the hidden states of the source sequence $\bm{x}_S$ and target sequence $\bm{x}_T$ are acquired by the multi-head attention
\begin{equation}
\bm{h}^{l+1}_{S}=\text{MultiHead}(\bm{h}^l_S,\bm{h}^l_S,\bm{h}^l_S)
,~
\bm{h}^{l+1}_{T}=\text{MultiHead}(\bm{h}^l_T,\bm{h}^l_T,\bm{h}^l_T).
\end{equation}

\textbf{Manifold Mixup}\quad
To reduce the cross-lingual representation discrepancy, a straightforward idea is to find compromised representations between the source and target languages.
It's difficult to find such representations because of varying degrees of differences across languages, like vocabulary and word order.
However, manifold mixup provides an elegant way to get intermediate representations by conducting linear interpolation on hidden states.

To extract target-related information from the source hidden states, the target hidden states are used as the query, and the source hidden states are used as the key and value. This cross-attention process is computed as
\begin{equation}
 \bm{h}^{l+1}_{T|S}=\text{MultiHead}(\bm{h}^{l+1}_T,\bm{h}^{l+1}_S,\bm{h}^{l+1}_S),
\end{equation}
which shares parameters with the multi-head attention.
The manifold mixup process mixes the target hidden states $\bm{h}^{l+1}_{T}$ and the source-aware target hidden states $\bm{h}^{l+1}_{T|S}$ based on the \textit{mixup ratio} $\lambda$
\begin{equation}
\tilde{\bm{h}}^{l+1}_T=\text{LN}(\lambda \bm{h}^{l+1}_{T|S}+(1-\lambda )\bm{h}^{l+1}_T),
\end{equation}
where $\lambda$ is an instance-level parameter, ranging from 0 to 1, and indicates the degree of manifold mixup. $\text{LN}$ denotes the layer norm operation.

\textbf{Mixup Ratio}\quad
The machine translation process may change the original semantics and introduce data noises in varying degrees \citep{Castilho2017,Fomicheva2020}.
Thus we introduce the translation quality modeling in the mixup process to handle this problem.
Following \citet{Fomicheva2020}, we use the entropy of attention weights to measure the translation quality
\begin{equation}\label{eq_10}
\text{H}(\bm{A})=-\frac{1}{I}\sum_i^I\sum_j^J \bm{A}_{ji}\text{log}\bm{A}_{ji}
\text{, where } \bm{A}_{ij}=\text{softmax}(\frac{\bm{h}_{T_i}\bm{h}_{S_j}^\top}{\sqrt{n}}).
\end{equation}
$I$ is the number of target tokens and $J$ is the number of source tokens. Lower entropy implies better cross-lingual alignment and higher translation quality.

To introduce the translation quality modeling into the manifold mixup process, we compute the mixup ratio as $\lambda = \lambda_0 \cdot \sigma[(\text{H}(\bm{A}) + \text{H}(\bm{A}^\top))W+b]$,
where $\sigma$ is the sigmoid function, and $W$, $b$ are trainable parameters. $\lambda_0$ is the max value of the mixup ratio, which is set to 0.5 in this paper. We consider two-way alignment in the translation quality modeling, i.e. $\text{H}(\bm{A})$ and $\text{H}(\bm{A}^\top)$.

\textbf{Scheduled Sampling}\quad
The source sequences utilized in training and inference are drawn from different distributions.
During training, the source sequence is a real text from $\mathcal{D}_S^{\text{Train}}$, while during inference, the source sequence is a translation from $\tilde{\mathcal{D}}_S^{\text{Test}}$. This discrepancy, commonly called the exposure bias, leads to a gap between training and inference.

Motivated by the scheduled sampling approach \citep{Bengio2015} in NMT, we sample the source sequence dynamically during training. Specifically, the source sequence fed into the manifold mixup is either a real text from $\mathcal{D}_S^{\text{Train}}$ or translation from $\tilde{\mathcal{D}}_S^{\text{Train}}$ with a certain probability $p$
\begin{equation}
\left\{
\begin{aligned}
p <=p^*, &      & \bm{x}_s \in \mathcal{D}_S^{\text{Train}},  \\
p > p^*,  &      & \bm{x}_s \in \tilde{\mathcal{D}}_S^{\text{Train}},
\end{aligned}
\right.
\end{equation}
where $p^*$ is decreasing during training to match the situation of inference.
We utilize the inverse sigmoid decay \citep{Bengio2015}, which decreases $p^*$ as a function of the index of mini-batch.

\subsection{Final Training Objective}
The training loss is composed of two parts: the task loss and the consistency loss
\begin{equation}
\mathcal{L}=\underbrace{\mathcal{L}_{\text{task}}}_{\text{task loss}}+ \underbrace{\text{MSE}(\bm{r}_S, \bm{r}_T) + \text{KL}(\bm{p}_S, \bm{p}_T)}_{\text{consistency loss}}.
\label{eq_loss}
\end{equation}
where $\text{MSE}(\cdot)$ is Mean Squared Error and $\text{KL}(\cdot)$ is Kullback-Leibler divergence.
$\bm{r}_*$ is the sequence representation\footnote{We utilize the mean pooling of the last layer's hidden states as the sequence representation, which is independent of the sequence length.}
and $\bm{p}_*$ is the predicted probability distribution of downstream tasks.

The task loss $\mathcal{L}_{\text{task}}$ is the sum of the source language task loss $\mathcal{L}_{\text{task}}^S$ and target language one $\mathcal{L}_{\text{task}}^T$,
weighted by the hyper-parameter $\alpha$, which is utilized to balance the training process
\begin{equation}
\label{final_task_loss}
\mathcal{L}_{\text{task}}=\alpha\mathcal{L}_{\text{task}}^S + (1-\alpha) \mathcal{L}_{\text{task}}^T.
\end{equation}
For classification, structured prediction, and span extraction tasks, the task loss is the cross-entropy loss (see details in Appendix \ref{task_loss}).
For structured prediction tasks, it is non-trivial to implement the token-level label mapping across different languages.
Thus we use the label probability distribution, predicted by the source language task model, as the pseudo-label for training, where tokens and labels are corresponding.

The consistency loss is composed of two parts: the representation consistency loss and the prediction consistency loss.
The first loss is a regularization term and provides a way to align representations across different languages \citep{Ruder2019}.
The second loss is to make better use of the supervision of downstream tasks. It only exists in the classification task, as the translation process does not change the label of this task, while in other tasks, it does.

\subsection{Inference}
During inference, the manifold mixup process is the same as training, except for the Scheduled Sampling process.
Concretely, for the source language, only translation data are available in the inference stage, without real data, so we use $\bm{x}_s \in \tilde{\mathcal{D}}_S^{\text{Test}}$.
For classification tasks, we synthesize the predictions of both the source and target sequences by taking the mean of the predicted probability distributions as the final prediction. For structured prediction and QA tasks, we only consider the prediction of the target sequence.

\section{Experiments}
This section first introduces the cross-lingual understanding benchmark, XTREME.
Then briefly introduces the configurations of downstream tasks and baselines. 
Finally, shows the main results of baselines and \method on XTREME.

\subsection{Tasks and Settings}
\textbf{Tasks}\quad
In our experiments, we focus on three types of tasks in XTREME:
(1) sentence pair classification task: XNLI \citep{xnli} and PAWS-X \citep{pawsx}; (2) structured prediction task: POS \citep{nivre:hal-01930733} and NER \citep{pan-etal-2017-cross}; (3) question answering task: XQuAD \citep{xquad}, MLQA \citep{mlqa} and TyDiQA \citep{tydiqa}.
The details of these datasets can refer to \citet{xtreme}.
We utilize the translate-train and translate-test data from the XTREME repo\footnote{\url{https://github.com/google-research/xtreme}.}, which also provide the pseudo-label of translate-train data for classification tasks and question answering tasks.
The rest translation data are from Google Translate\footnote{\url{https://translate.google.com/}.}.

\textbf{Models}\quad
Experiments are based on two multilingual pre-trained models: mBERT and XLM-R. We use the pre-trained models of Huggingface Transformers\footnote{We use \texttt{bert-base-multilingual-cased} for mBERT
and \texttt{xlm-roberta-large} for XLM-R.} as the backbone model.

\textbf{Hyper-parameters}\quad
We select XNLI, POS, and MLQA as representative tasks to search for the best hyper-parameters.
The final model is selected based on the averaged performance of all languages on the dev set.
We perform grid search over the balance training parameter $\alpha$ and learning rate from [0.2, 0.4, 0.6, 0.8] and [3e-6, 5e-6, 2e-5, 3e-5].
We also search for the best manifold mixup layer from [1, 4, 8, 12, 16, 20, 24]. In final results, we implement mixup in the first layer for classification tasks, 4th layer for structured prediction tasks. For QA tasks, we implement mixup in the 16th layer for large model, 8th layer for base model.
Concrete details of experiments are presented in Appendix \ref{hyper}.

\begin{table*}[t!]
\centering
\setlength{\belowcaptionskip}{0.2cm}
\caption{Main results on the XTREME benchmark. 
$^{\dagger}$ denotes using other data augmentation strategy in addition to machine translation. 
$^{\ddagger}$ denotes results from \citet{xtreme-r}, which is an updated version of \citet{xtreme}.
}
\resizebox{\linewidth}{!}{
\begin{tabular}{l c c c c c c c c}
\toprule
\multirow{2}{*}{Model} & \multicolumn{2}{c}{Pair Sentence} & \multicolumn{2}{c}{Structured Prediction} & \multicolumn{3}{c}{Question Answering} & \multirow{2}{*}{Avg.} \\
 & XNLI & PAWS-X & POS & NER & XQuAD & MLQA & TyDiQA & \\
\midrule
Metrics & Acc & Acc & F1 & F1 & F1/EM & F1/EM & F1/EM & - \\
\midrule
\textsl{Based on XLM-R-large} & & & & & & & & \\
XLM-R \citep{xtreme} & 79.2 & 86.4 & 73.8 & 65.4 & 76.6/60.8 & 71.6/53.2 & 65.1/45.0 & 70.1 \\
Trans-train \citep{hictl} & 82.9 & 90.1 & 74.6 & 66.8 & 80.4/65.6 & 72.4/54.7 & 66.2/48.2 & 72.6 \\
Filter \citep{Filter} & 83.9 & 91.4 & 76.2 & 67.7 & 82.4/68.0 & 76.2/57.7 & 68.3/50.9 & 74.4 \\
\textsc{xTune} \citep{zheng2021} & 84.8 & 91.6 & \textbf{79.3}$^{\dagger}$ & \textbf{69.9}$^{\dagger}$ & 82.5/69.0$^{\dagger}$ & 75.0/57.1$^{\dagger}$ & \textbf{75.4/60.8}$^{\dagger}$ & \textbf{76.5} \\ 
\method & \textbf{85.3} & \textbf{91.8} & 78.4 & 69.0 & \textbf{82.6/69.3} & \textbf{76.5/58.1} & 69.0/52.8 & 75.5 \\
\midrule
\textsl{Based on mBERT} & & & & & & & & \\
mBERT \citep{xtreme} & 65.4 & 81.9 & 71.5 & 62.2 & 64.5/49.4 & 61.4/44.2 & 59.7/43.9 & 63.2 \\
Joint-Align \citep{zhao2020} & 72.3 & - & - & - & - & - & - & - \\
Trans-train \citep{xtreme} & 75.1 & 88.9 & - & - & 72.4/58.3 & 67.6/49.8 & 59.5/45.8$^{\ddagger}$ & - \\
\method & \textbf{78.8} & \textbf{89.7} & \textbf{76.5} & \textbf{65.0} & \textbf{73.3/58.9} & \textbf{69.0/50.9} & \textbf{60.8/46.5} & \textbf{70.0} \\
\bottomrule
\end{tabular}
}
\label{table_main}
\end{table*}
\subsection{Baselines}
We conduct experiments on two strong multilingual pre-trained models to verify the generality of methods:
(1) \textbf{mBERT}\quad Multilingual BERT is a 12-layer transformer model pre-trained on the Wikipedias of 104 languages.
(2) \textbf{XLM-R}\quad XLM-R-large is a 24-layer transformer model pre-trained on 2.5T data extracted from Common Crawl covering 100 languages.
Based on them, these are some strong baselines:
(1) \textbf{Trans-train}\quad Abbreviation for Translate-train. The training set of the source language is machine-translated to each target language and then the model is trained on the concatenation of all training sets.
(2) \textbf{Joint-Align}\quad
\citet{zhao2020} aligns the monolingual sub-spaces of the source and target language by minimizing the distances of embeddings for matched word pairs.
(3) \textbf{Filter}\quad
\citet{Filter} splices the representation of the target sequence and its translation in intermediate layers to extract multilingual knowledge.
(4) \textbf{\textsc{xTune}}\quad
\citet{zheng2021} uses two types of consistency regularization based on four types of data augmentation.
\begin{table}[t]
\footnotesize
\setlength{\belowcaptionskip}{0.2cm}
\centering
\caption{Comparisons between \method and \textsc{xTune} under the same setting: XLM-R-base model and machine translation data augmentation. Results of \textsc{xTune} are from \citet{zheng2021} Table 4.}
\begin{tabular}{l c c c}
\toprule
Model & XNLI & POS & MLQA \\\midrule
\textsc{xTune}$_{\mathcal{R}_1}$ \citep{zheng2021} & 79.7 & - & - \\ 
\textsc{xTune}$_{\mathcal{R}_2}$ \citep{zheng2021} & 78.9 & 76.6 & 68.7/51.1 \\
\method & \textbf{80.4} & \textbf{77.8} & \textbf{71.2/53.1} \\
\bottomrule
\end{tabular}
\label{table_xtune}
\end{table}

\subsection{Main Results}
Results on the XTREME benchmark are shown in Table \ref{table_main}.
Concrete results for each task are presented in Appendix \ref{detail_result}.
Compared with strong baselines, \method shows its superiority across different backbones and tasks, which indicates its generality.
\method outperforms Trans-train by 2.2\% based on mBERT, and \method outperforms Filter by 1.5\% based on XLM-R.
The superiority of \method over Filter is that \method gives a calibrated representation for target languages, not just the concatenation of two representations.
Besides, \method considers the noise of translation data and limits the noise propagation by introducing mixup ratio.

\textsc{xTune} achieves the best results on structured prediction tasks and the low-resource QA task TyDiQA (only 3.7k training data in English), but \textsc{xTune} uses three other data augmentation approaches in addition to machine translation.
To make a fairer comparison, we conduct experiments under the same setting in Table \ref{table_xtune}, which indicates \method outperforms \textsc{xTune} on three types of tasks with only machine translation data augmentation.
Besides, \method only needs one-stage training, while \textsc{xTune} implements a two-stage training algorithm.
However, \method and \textsc{xTune} are complementary, where the former focuses on finding better representations for target languages while the latter concentrates on the cross-lingual data augmentation and consistency regularization.

\section{Analysis and Discussion}
To better understand \method and explore how \method influences the cross-lingual transfer performance, we conduct analyses\footnote{In this section, we utilize the XLM-R-large model as the backbone model.} on several questions.
Results show \method achieves performance improvements across different languages and it also reduces the cross-lingual representation discrepancy obviously. Table \ref{table_xnli_new_lang} in Appendix \ref{appendix_analysis} verifies the effectiveness of X-Mixup on both seen and unseen languages.
Besides, ablation results show the cross-lingual manifold mixup training contributes a lot to cross-lingual transfer. 

\begin{table}[t]
\footnotesize
\setlength{\belowcaptionskip}{0.2cm}
\centering
\caption{Ablation results on \method, where w/o mixup denotes remove the cross-lingual manifold mixup during training and inference and $\lambda=\lambda_0$ denotes a constant mixup ratio.}
\begin{tabular}{l c c c}
\toprule
Model & XNLI & POS & MLQA \\
\midrule
\method & 85.3 & 78.4 & 76.5/58.1 \\
\quad w/o mixup & 82.9 & 75.7 & 72.7/54.8 \\
\quad w/o mixup inference & 84.0 & 77.6 & 75.6/57.3 \\
\quad w/o scheduled sampling & 84.6 & 78.0 & 76.3/57.9 \\
\quad w/o consistency loss & 84.2 & 78.0 & 76.5/58.0 \\
\quad $\lambda=\lambda_0$ & 84.1 & 77.8 & 75.8/57.5 \\
\bottomrule
\end{tabular}
\label{table_ablation}
\end{table}

\noindent \textbf{(Q1)} \textbf{\textit{How \method influences the cross-lingual representation discrepancy?}}
Language centroid \citep{rosenberg2007} is the mean of the representations within each language.
We plot the language centroid of different methods (see Figure \ref{fig_cka_rep} in Appendix \ref{appendix_analysis}), which indicates \method brings closer language centroids significantly.
We also calculate the CKA scores of the XNLI dataset (see Table \ref{table_xnli_cka} in Appendix \ref{appendix_analysis}).
Results show \method reduces the cross-lingual representation discrepancy evenly across different target languages, improving the CKA score by 10.4\% on average.
In conclusion, both the language centroids visualization and the CKA score improvement indicate \method reduces the cross-lingual representation discrepancy effectively.

\noindent \textbf{(Q2)} \textbf{\textit{How \method influences the cross-lingual transfer gap?}}
We compare the cross-lingual transfer gap in Appendix \ref{appendix_analysis} Table \ref{table_gap}.
Compared with Trans-train, \method reduces the averaged gap by 39.8\% and shows its superiority across three types of tasks.
Compared with state-of-the-art methods, \method achieves the smallest cross-lingual transfer gap on four out of seven datasets, which suggests the effectiveness of \method on classification and QA tasks.

\noindent \textbf{(Q3)} \textbf{\textit{What is the essential component of \method?}}
There are five major components of \method: cross-lingual manifold mixup training, mixup inference, Mixup Ratio, Scheduled Sampling, and consistency loss.
To better understand \method, we implement ablation studies in Table \ref{table_ablation}.
Comparisons between \method and w/o mixup show the effectiveness of cross-lingual manifold mixup across different tasks, and even without mixup inference (translate-test data), the mixup training can also achieve 2.6\% improvements on average.
Besides, comparisons between \method and $\lambda=\lambda_0$ show the effectiveness of introducing the translation quality modeling in the mixup process.
Scheduled sampling achieves more performance improvements on the classification task, as the task shares labels across languages, and scheduled sampling can prevent the model from solely relying on the gold source sequence to make predictions.
In addition, the consistency loss is also more effective on the classification task, because there is additional prediction consistency loss which can transfer the task capability from the source language to target languages.
Detailed ablation results on the consistency loss are shown in Appendix \ref{appendix_analysis} Table \ref{table_ablation_consist}, which shows the KL consistency loss contributes more than the MSE consistency loss on the classification task.

\noindent \textbf{(Q4)} \textbf{\textit{Which layer is the best to implement the manifold mixup?}}
We implement the cross-lingual manifold mixup in different layers (see Figure \ref{fig_layer} in Appendix \ref{appendix_analysis} for details) and find different tasks prefer different mixup layers.
Although different tasks have their own preferences, no matter which layer we mix, the cross-lingual transfer performance can be improved, except for mixing within a higher layer on classification tasks.
The drop in classification task is mainly because the source and target sequences share the same task label. Performing mixup in a higher layer may make the model rely on the source sequence and ignore the target sequence.
The structured prediction task is not sensitive to the mixup layer, mainly because this task relies on both the short and long dependence.
For QA tasks, the cross-lingual transfer performance shows a trend from rise to decline as the mixup layer increases. The QA task needs higher-level understanding, but higher layers are more language-specific, where sequences from different languages have different gold answers.

\section{Conclusion}
This paper focuses on enhancing the cross-lingual transfer performance on understanding tasks.
Considering the large cross-lingual transfer gap in recent works, this paper first analyses related factors and finds this gap is strongly associated with the cross-lingual representation discrepancy.
Then \method is proposed to alleviate the discrepancy, which gives compromised representations for target languages by implementing the manifold mixup between the source and target languages.
Empirical evaluations on XTREME verify the effectiveness of \method across different tasks and languages. Besides, both the visualization and quantitative analyses show \method reduces the cross-lingual representation discrepancy effectively.
Furthermore, X-Mixup can also be applied to the multilingual pre-training process by implementing the cross-lingual manifold mixup on parallel data.
Findings on the relationship between the cross-lingual transfer performance and representation discrepancy shed light on a promising way to boost cross-lingual transfer for future research.

\bibliography{iclr2022_conference}

\begin{thebibliography}{52}
\providecommand{\natexlab}[1]{#1}
\providecommand{\url}[1]{\texttt{#1}}
\expandafter\ifx\csname urlstyle\endcsname\relax
  \providecommand{\doi}[1]{doi: #1}\else
  \providecommand{\doi}{doi: \begingroup \urlstyle{rm}\Url}\fi

\bibitem[Artetxe et~al.(2020)Artetxe, Ruder, and Yogatama]{xquad}
Mikel Artetxe, Sebastian Ruder, and Dani Yogatama.
\newblock On the cross-lingual transferability of monolingual representations.
\newblock In \emph{Proceedings of the 58th Annual Meeting of the Association
  for Computational Linguistics}, pp.\  4623--4637, Online, July 2020.
  Association for Computational Linguistics.
\newblock \doi{10.18653/v1/2020.acl-main.421}.
\newblock URL \url{https://www.aclweb.org/anthology/2020.acl-main.421}.

\bibitem[Bengio et~al.(2015)Bengio, Vinyals, Jaitly, and Shazeer]{Bengio2015}
Samy Bengio, Oriol Vinyals, Navdeep Jaitly, and Noam Shazeer.
\newblock Scheduled sampling for sequence prediction with recurrent neural
  networks.
\newblock In Corinna Cortes, Neil~D. Lawrence, Daniel~D. Lee, Masashi Sugiyama,
  and Roman Garnett (eds.), \emph{Advances in Neural Information Processing
  Systems 28: Annual Conference on Neural Information Processing Systems 2015,
  December 7-12, 2015, Montreal, Quebec, Canada}, pp.\  1171--1179, 2015.
\newblock URL
  \url{https://proceedings.neurips.cc/paper/2015/hash/e995f98d56967d946471af29d7bf99f1-Abstract.html}.

\bibitem[Bornea et~al.(2020)Bornea, Pan, Rosenthal, Florian, and
  Sil]{Mihaela2020}
Mihaela Bornea, Lin Pan, Sara Rosenthal, Radu Florian, and Avirup Sil.
\newblock Multilingual transfer learning for {QA} using translation as data
  augmentation.
\newblock \emph{CoRR}, abs/2012.05958, 2020.
\newblock URL \url{https://arxiv.org/abs/2012.05958}.

\bibitem[Castilho et~al.(2017)Castilho, Moorkens, Gaspari, Calixto, Tinsley,
  and Way]{Castilho2017}
Sheila Castilho, Joss Moorkens, Federico Gaspari, Iacer Calixto, John Tinsley,
  and Andy Way.
\newblock Is neural machine translation the new state of the art?
\newblock \emph{Prague Bull. Math. Linguistics}, 108:\penalty0 109--120, 2017.
\newblock URL
  \url{http://ufal.mff.cuni.cz/pbml/108/art-castilho-moorkens-gaspari-tinsley-calixto-way.pdf}.

\bibitem[Chen et~al.(2020)Chen, Yang, and Yang]{chen-etal-2020-mixtext}
Jiaao Chen, Zichao Yang, and Diyi Yang.
\newblock {M}ix{T}ext: Linguistically-informed interpolation of hidden space
  for semi-supervised text classification.
\newblock In \emph{Proceedings of the 58th Annual Meeting of the Association
  for Computational Linguistics}, pp.\  2147--2157, Online, July 2020.
  Association for Computational Linguistics.
\newblock \doi{10.18653/v1/2020.acl-main.194}.
\newblock URL \url{https://aclanthology.org/2020.acl-main.194}.

\bibitem[Chen et~al.(2019)Chen, Awadallah, Hassan, Wang, and Cardie]{Chen2019}
Xilun Chen, Ahmed~Hassan Awadallah, Hany Hassan, Wei Wang, and Claire Cardie.
\newblock Multi-source cross-lingual model transfer: Learning what to share.
\newblock In \emph{Proceedings of the 57th Annual Meeting of the Association
  for Computational Linguistics}, pp.\  3098--3112, Florence, Italy, July 2019.
  Association for Computational Linguistics.
\newblock \doi{10.18653/v1/P19-1299}.
\newblock URL \url{https://www.aclweb.org/anthology/P19-1299}.

\bibitem[Chi et~al.(2020)Chi, Dong, Wei, Yang, Singhal, Wang, Song, Mao, Huang,
  and Zhou]{infoxlm}
Zewen Chi, Li~Dong, Furu Wei, Nan Yang, Saksham Singhal, Wenhui Wang, Xia Song,
  Xian{-}Ling Mao, Heyan Huang, and Ming Zhou.
\newblock Infoxlm: An information-theoretic framework for cross-lingual
  language model pre-training.
\newblock \emph{CoRR}, abs/2007.07834, 2020.
\newblock URL \url{https://arxiv.org/abs/2007.07834}.

\bibitem[Clark et~al.(2020)Clark, Choi, Collins, Garrette, Kwiatkowski,
  Nikolaev, and Palomaki]{tydiqa}
Jonathan~H. Clark, Eunsol Choi, Michael Collins, Dan Garrette, Tom Kwiatkowski,
  Vitaly Nikolaev, and Jennimaria Palomaki.
\newblock {T}y{D}i {QA}: A benchmark for information-seeking question answering
  in typologically diverse languages.
\newblock \emph{Transactions of the Association for Computational Linguistics},
  8:\penalty0 454--470, 2020.
\newblock \doi{10.1162/tacl_a_00317}.
\newblock URL \url{https://www.aclweb.org/anthology/2020.tacl-1.30}.

\bibitem[Conneau \& Lample(2019)Conneau and Lample]{xlm}
Alexis Conneau and Guillaume Lample.
\newblock Cross-lingual language model pretraining.
\newblock In Hanna~M. Wallach, Hugo Larochelle, Alina Beygelzimer, Florence
  d'Alch{\'{e}}{-}Buc, Emily~B. Fox, and Roman Garnett (eds.), \emph{Advances
  in Neural Information Processing Systems 32: Annual Conference on Neural
  Information Processing Systems 2019, NeurIPS 2019, December 8-14, 2019,
  Vancouver, BC, Canada}, pp.\  7057--7067, 2019.
\newblock URL
  \url{https://proceedings.neurips.cc/paper/2019/hash/c04c19c2c2474dbf5f7ac4372c5b9af1-Abstract.html}.

\bibitem[Conneau et~al.(2018)Conneau, Rinott, Lample, Williams, Bowman,
  Schwenk, and Stoyanov]{xnli}
Alexis Conneau, Ruty Rinott, Guillaume Lample, Adina Williams, Samuel Bowman,
  Holger Schwenk, and Veselin Stoyanov.
\newblock {XNLI}: Evaluating cross-lingual sentence representations.
\newblock In \emph{Proceedings of the 2018 Conference on Empirical Methods in
  Natural Language Processing}, pp.\  2475--2485, Brussels, Belgium,
  October-November 2018. Association for Computational Linguistics.
\newblock \doi{10.18653/v1/D18-1269}.
\newblock URL \url{https://www.aclweb.org/anthology/D18-1269}.

\bibitem[Conneau et~al.(2020{\natexlab{a}})Conneau, Khandelwal, Goyal,
  Chaudhary, Wenzek, Guzm{\'a}n, Grave, Ott, Zettlemoyer, and Stoyanov]{xlmr}
Alexis Conneau, Kartikay Khandelwal, Naman Goyal, Vishrav Chaudhary, Guillaume
  Wenzek, Francisco Guzm{\'a}n, Edouard Grave, Myle Ott, Luke Zettlemoyer, and
  Veselin Stoyanov.
\newblock Unsupervised cross-lingual representation learning at scale.
\newblock In \emph{Proceedings of the 58th Annual Meeting of the Association
  for Computational Linguistics}, pp.\  8440--8451, Online, July
  2020{\natexlab{a}}. Association for Computational Linguistics.
\newblock \doi{10.18653/v1/2020.acl-main.747}.
\newblock URL \url{https://www.aclweb.org/anthology/2020.acl-main.747}.

\bibitem[Conneau et~al.(2020{\natexlab{b}})Conneau, Wu, Li, Zettlemoyer, and
  Stoyanov]{conneau2020}
Alexis Conneau, Shijie Wu, Haoran Li, Luke Zettlemoyer, and Veselin Stoyanov.
\newblock Emerging cross-lingual structure in pretrained language models.
\newblock In \emph{Proceedings of the 58th Annual Meeting of the Association
  for Computational Linguistics}, pp.\  6022--6034, Online, July
  2020{\natexlab{b}}. Association for Computational Linguistics.
\newblock \doi{10.18653/v1/2020.acl-main.536}.
\newblock URL \url{https://www.aclweb.org/anthology/2020.acl-main.536}.

\bibitem[Devlin et~al.(2019)Devlin, Chang, Lee, and Toutanova]{bert}
Jacob Devlin, Ming-Wei Chang, Kenton Lee, and Kristina Toutanova.
\newblock {BERT}: Pre-training of deep bidirectional transformers for language
  understanding.
\newblock In \emph{Proceedings of the 2019 Conference of the North {A}merican
  Chapter of the Association for Computational Linguistics: Human Language
  Technologies, Volume 1 (Long and Short Papers)}, pp.\  4171--4186,
  Minneapolis, Minnesota, June 2019. Association for Computational Linguistics.
\newblock \doi{10.18653/v1/N19-1423}.
\newblock URL \url{https://www.aclweb.org/anthology/N19-1423}.

\bibitem[Fang et~al.(2020)Fang, Wang, Gan, Sun, and Liu]{Filter}
Yuwei Fang, Shuohang Wang, Zhe Gan, Siqi Sun, and Jingjing Liu.
\newblock {FILTER:} an enhanced fusion method for cross-lingual language
  understanding.
\newblock \emph{CoRR}, abs/2009.05166, 2020.
\newblock URL \url{https://arxiv.org/abs/2009.05166}.

\bibitem[Fomicheva et~al.(2020)Fomicheva, Sun, Yankovskaya, Blain,
  Guzm{\'{a}}n, Fishel, Aletras, Chaudhary, and Specia]{Fomicheva2020}
Marina Fomicheva, Shuo Sun, Lisa Yankovskaya, Fr{\'{e}}d{\'{e}}ric Blain,
  Francisco Guzm{\'{a}}n, Mark Fishel, Nikolaos Aletras, Vishrav Chaudhary, and
  Lucia Specia.
\newblock Unsupervised quality estimation for neural machine translation.
\newblock \emph{Trans. Assoc. Comput. Linguistics}, 8:\penalty0 539--555, 2020.
\newblock URL \url{https://transacl.org/ojs/index.php/tacl/article/view/1997}.

\bibitem[Hu et~al.(2020)Hu, Ruder, Siddhant, Neubig, Firat, and
  Johnson]{xtreme}
Junjie Hu, Sebastian Ruder, Aditya Siddhant, Graham Neubig, Orhan Firat, and
  Melvin Johnson.
\newblock Xtreme: A massively multilingual multi-task benchmark for evaluating
  cross-lingual generalization.
\newblock \emph{CoRR}, abs/2003.11080, 2020.

\bibitem[Hu et~al.(2021)Hu, Johnson, Firat, Siddhant, and Neubig]{amber}
Junjie Hu, Melvin Johnson, Orhan Firat, Aditya Siddhant, and Graham Neubig.
\newblock Explicit alignment objectives for multilingual bidirectional
  encoders.
\newblock In Kristina Toutanova, Anna Rumshisky, Luke Zettlemoyer, Dilek
  Hakkani{-}T{\"{u}}r, Iz~Beltagy, Steven Bethard, Ryan Cotterell, Tanmoy
  Chakraborty, and Yichao Zhou (eds.), \emph{Proceedings of the 2021 Conference
  of the North American Chapter of the Association for Computational
  Linguistics: Human Language Technologies, {NAACL-HLT} 2021, Online, June
  6-11, 2021}, pp.\  3633--3643. Association for Computational Linguistics,
  2021.
\newblock URL \url{https://doi.org/10.18653/v1/2021.naacl-main.284}.

\bibitem[Huang et~al.(2019)Huang, Liang, Duan, Gong, Shou, Jiang, and
  Zhou]{unicoder}
Haoyang Huang, Yaobo Liang, Nan Duan, Ming Gong, Linjun Shou, Daxin Jiang, and
  Ming Zhou.
\newblock {U}nicoder: A universal language encoder by pre-training with
  multiple cross-lingual tasks.
\newblock In \emph{Proceedings of the 2019 Conference on Empirical Methods in
  Natural Language Processing and the 9th International Joint Conference on
  Natural Language Processing (EMNLP-IJCNLP)}, pp.\  2485--2494, Hong Kong,
  China, November 2019. Association for Computational Linguistics.
\newblock \doi{10.18653/v1/D19-1252}.
\newblock URL \url{https://www.aclweb.org/anthology/D19-1252}.

\bibitem[Jindal et~al.(2020)Jindal, Chowdhury, Didolkar, Jin, Sawhney, and
  Shah]{Jindal2020}
Amit Jindal, Arijit~Ghosh Chowdhury, Aniket Didolkar, Di~Jin, Ramit Sawhney,
  and Rajiv~Ratn Shah.
\newblock Augmenting {NLP} models using latent feature interpolations.
\newblock In Donia Scott, N{\'{u}}ria Bel, and Chengqing Zong (eds.),
  \emph{Proceedings of the 28th International Conference on Computational
  Linguistics, {COLING} 2020, Barcelona, Spain (Online), December 8-13, 2020},
  pp.\  6931--6936. International Committee on Computational Linguistics, 2020.
\newblock \doi{10.18653/v1/2020.coling-main.611}.
\newblock URL \url{https://doi.org/10.18653/v1/2020.coling-main.611}.

\bibitem[Joshi et~al.(2020)Joshi, Santy, Budhiraja, Bali, and
  Choudhury]{Joshi2020}
Pratik Joshi, Sebastin Santy, Amar Budhiraja, Kalika Bali, and Monojit
  Choudhury.
\newblock The state and fate of linguistic diversity and inclusion in the {NLP}
  world.
\newblock In \emph{Proceedings of the 58th Annual Meeting of the Association
  for Computational Linguistics}, pp.\  6282--6293, Online, July 2020.
  Association for Computational Linguistics.
\newblock \doi{10.18653/v1/2020.acl-main.560}.
\newblock URL \url{https://www.aclweb.org/anthology/2020.acl-main.560}.

\bibitem[Kale et~al.(2021)Kale, Siddhant, Al{-}Rfou, Xue, Constant, and
  Johnson]{nmt5}
Mihir Kale, Aditya Siddhant, Rami Al{-}Rfou, Linting Xue, Noah Constant, and
  Melvin Johnson.
\newblock nmt5 - is parallel data still relevant for pre-training massively
  multilingual language models?
\newblock In Chengqing Zong, Fei Xia, Wenjie Li, and Roberto Navigli (eds.),
  \emph{Proceedings of the 59th Annual Meeting of the Association for
  Computational Linguistics and the 11th International Joint Conference on
  Natural Language Processing, {ACL/IJCNLP} 2021, (Volume 2: Short Papers),
  Virtual Event, August 1-6, 2021}, pp.\  683--691. Association for
  Computational Linguistics, 2021.
\newblock \doi{10.18653/v1/2021.acl-short.87}.
\newblock URL \url{https://doi.org/10.18653/v1/2021.acl-short.87}.

\bibitem[Keung et~al.(2019)Keung, Lu, and Bhardwaj]{keung2019}
Phillip Keung, Yichao Lu, and Vikas Bhardwaj.
\newblock Adversarial learning with contextual embeddings for zero-resource
  cross-lingual classification and {NER}.
\newblock In \emph{Proceedings of the 2019 Conference on Empirical Methods in
  Natural Language Processing and the 9th International Joint Conference on
  Natural Language Processing (EMNLP-IJCNLP)}, pp.\  1355--1360, Hong Kong,
  China, November 2019. Association for Computational Linguistics.
\newblock \doi{10.18653/v1/D19-1138}.
\newblock URL \url{https://www.aclweb.org/anthology/D19-1138}.

\bibitem[Kornblith et~al.(2019)Kornblith, Norouzi, Lee, and
  Hinton]{Kornblith2019}
Simon Kornblith, Mohammad Norouzi, Honglak Lee, and Geoffrey~E. Hinton.
\newblock Similarity of neural network representations revisited.
\newblock In Kamalika Chaudhuri and Ruslan Salakhutdinov (eds.),
  \emph{Proceedings of the 36th International Conference on Machine Learning,
  {ICML} 2019, 9-15 June 2019, Long Beach, California, {USA}}, volume~97 of
  \emph{Proceedings of Machine Learning Research}, pp.\  3519--3529. {PMLR},
  2019.
\newblock URL \url{http://proceedings.mlr.press/v97/kornblith19a.html}.

\bibitem[Lauscher et~al.(2020)Lauscher, Ravishankar, Vuli{\'c}, and
  Glava{\v{s}}]{Lauscher2020}
Anne Lauscher, Vinit Ravishankar, Ivan Vuli{\'c}, and Goran Glava{\v{s}}.
\newblock From zero to hero: {O}n the limitations of zero-shot language
  transfer with multilingual {T}ransformers.
\newblock In \emph{Proceedings of the 2020 Conference on Empirical Methods in
  Natural Language Processing (EMNLP)}, pp.\  4483--4499, Online, November
  2020. Association for Computational Linguistics.
\newblock \doi{10.18653/v1/2020.emnlp-main.363}.
\newblock URL \url{https://www.aclweb.org/anthology/2020.emnlp-main.363}.

\bibitem[Lewis et~al.(2020)Lewis, Oguz, Rinott, Riedel, and Schwenk]{mlqa}
Patrick Lewis, Barlas Oguz, Ruty Rinott, Sebastian Riedel, and Holger Schwenk.
\newblock {MLQA}: Evaluating cross-lingual extractive question answering.
\newblock In \emph{Proceedings of the 58th Annual Meeting of the Association
  for Computational Linguistics}, pp.\  7315--7330, Online, July 2020.
  Association for Computational Linguistics.
\newblock \doi{10.18653/v1/2020.acl-main.653}.
\newblock URL \url{https://www.aclweb.org/anthology/2020.acl-main.653}.

\bibitem[Libovick{\'y} et~al.(2020)Libovick{\'y}, Rosa, and
  Fraser]{libovicky2020}
Jind{\v{r}}ich Libovick{\'y}, Rudolf Rosa, and Alexander Fraser.
\newblock On the language neutrality of pre-trained multilingual
  representations.
\newblock In \emph{Findings of the Association for Computational Linguistics:
  EMNLP 2020}, pp.\  1663--1674, Online, November 2020. Association for
  Computational Linguistics.
\newblock \doi{10.18653/v1/2020.findings-emnlp.150}.
\newblock URL \url{https://www.aclweb.org/anthology/2020.findings-emnlp.150}.

\bibitem[Luo et~al.(2021)Luo, Wang, Liu, Liu, Bi, Huang, Huang, and Si]{veco}
Fuli Luo, Wei Wang, Jiahao Liu, Yijia Liu, Bin Bi, Songfang Huang, Fei Huang,
  and Luo Si.
\newblock {VECO:} variable and flexible cross-lingual pre-training for language
  understanding and generation.
\newblock In Chengqing Zong, Fei Xia, Wenjie Li, and Roberto Navigli (eds.),
  \emph{Proceedings of the 59th Annual Meeting of the Association for
  Computational Linguistics and the 11th International Joint Conference on
  Natural Language Processing, {ACL/IJCNLP} 2021, (Volume 1: Long Papers),
  Virtual Event, August 1-6, 2021}, pp.\  3980--3994. Association for
  Computational Linguistics, 2021.
\newblock \doi{10.18653/v1/2021.acl-long.308}.
\newblock URL \url{https://doi.org/10.18653/v1/2021.acl-long.308}.

\bibitem[Nivre et~al.(2018)Nivre, Abrams, Agi{c}, Ahrenberg, Antonsen, and
  et~al]{nivre:hal-01930733}
Joakim Nivre, Mitchell Abrams, {Z}eljko Agi{c}, Lars Ahrenberg, Lene Antonsen,
  and et~al.
\newblock {Universal Dependencies 2.2}, 2018.
\newblock URL \url{https://hal.archives-ouvertes.fr/hal-01930733}.
\newblock LINDAT/CLARIN digital library at the Institute of Formal and Applied
  Linguistics ({\'U}FAL), Faculty of Mathematics and Physics, Charles
  University.

\bibitem[Ouyang et~al.(2020)Ouyang, Wang, Pang, Sun, Tian, Wu, and
  Wang]{erniem}
Xuan Ouyang, Shuohuan Wang, Chao Pang, Yu~Sun, Hao Tian, Hua Wu, and Haifeng
  Wang.
\newblock {ERNIE-M:} enhanced multilingual representation by aligning
  cross-lingual semantics with monolingual corpora.
\newblock \emph{CoRR}, abs/2012.15674, 2020.
\newblock URL \url{https://arxiv.org/abs/2012.15674}.

\bibitem[Pan \& Yang(2010)Pan and Yang]{Pan2010}
Sinno~Jialin Pan and Qiang Yang.
\newblock A survey on transfer learning.
\newblock \emph{{IEEE} Trans. Knowl. Data Eng.}, 22\penalty0 (10):\penalty0
  1345--1359, 2010.
\newblock \doi{10.1109/TKDE.2009.191}.
\newblock URL \url{https://doi.org/10.1109/TKDE.2009.191}.

\bibitem[Pan et~al.(2017)Pan, Zhang, May, Nothman, Knight, and
  Ji]{pan-etal-2017-cross}
Xiaoman Pan, Boliang Zhang, Jonathan May, Joel Nothman, Kevin Knight, and Heng
  Ji.
\newblock Cross-lingual name tagging and linking for 282 languages.
\newblock In \emph{Proceedings of the 55th Annual Meeting of the Association
  for Computational Linguistics (Volume 1: Long Papers)}, pp.\  1946--1958,
  Vancouver, Canada, July 2017. Association for Computational Linguistics.
\newblock \doi{10.18653/v1/P17-1178}.
\newblock URL \url{https://www.aclweb.org/anthology/P17-1178}.

\bibitem[Ponti et~al.(2019)Ponti, O{'}Horan, Berzak, Vuli{\'c}, Reichart,
  Poibeau, Shutova, and Korhonen]{Ponti2019}
Edoardo~Maria Ponti, Helen O{'}Horan, Yevgeni Berzak, Ivan Vuli{\'c}, Roi
  Reichart, Thierry Poibeau, Ekaterina Shutova, and Anna Korhonen.
\newblock Modeling language variation and universals: A survey on typological
  linguistics for natural language processing.
\newblock \emph{Computational Linguistics}, 45\penalty0 (3):\penalty0 559--601,
  September 2019.
\newblock \doi{10.1162/coli_a_00357}.
\newblock URL \url{https://www.aclweb.org/anthology/J19-3005}.

\bibitem[Prettenhofer \& Stein(2011)Prettenhofer and Stein]{PrettenhoferS2011}
Peter Prettenhofer and Benno Stein.
\newblock Cross-lingual adaptation using structural correspondence learning.
\newblock \emph{{ACM} Trans. Intell. Syst. Technol.}, 3\penalty0 (1):\penalty0
  13:1--13:22, 2011.
\newblock \doi{10.1145/2036264.2036277}.
\newblock URL \url{https://doi.org/10.1145/2036264.2036277}.

\bibitem[Qin et~al.(2020)Qin, Ni, Zhang, and Che]{Qin2020}
Libo Qin, Minheng Ni, Yue Zhang, and Wanxiang Che.
\newblock Cosda-ml: Multi-lingual code-switching data augmentation for
  zero-shot cross-lingual {NLP}.
\newblock In Christian Bessiere (ed.), \emph{Proceedings of the Twenty-Ninth
  International Joint Conference on Artificial Intelligence, {IJCAI} 2020},
  pp.\  3853--3860. ijcai.org, 2020.
\newblock \doi{10.24963/ijcai.2020/533}.
\newblock URL \url{https://doi.org/10.24963/ijcai.2020/533}.

\bibitem[Ranzato et~al.(2016)Ranzato, Chopra, Auli, and Zaremba]{Ranzato2016}
Marc'Aurelio Ranzato, Sumit Chopra, Michael Auli, and Wojciech Zaremba.
\newblock Sequence level training with recurrent neural networks.
\newblock In Yoshua Bengio and Yann LeCun (eds.), \emph{4th International
  Conference on Learning Representations, {ICLR} 2016, San Juan, Puerto Rico,
  May 2-4, 2016, Conference Track Proceedings}, 2016.
\newblock URL \url{http://arxiv.org/abs/1511.06732}.

\bibitem[Rosenberg \& Hirschberg(2007)Rosenberg and Hirschberg]{rosenberg2007}
Andrew Rosenberg and Julia Hirschberg.
\newblock {V}-measure: A conditional entropy-based external cluster evaluation
  measure.
\newblock In \emph{Proceedings of the 2007 Joint Conference on Empirical
  Methods in Natural Language Processing and Computational Natural Language
  Learning ({EMNLP}-{C}o{NLL})}, pp.\  410--420, Prague, Czech Republic, June
  2007. Association for Computational Linguistics.
\newblock URL \url{https://www.aclweb.org/anthology/D07-1043}.

\bibitem[Ruder et~al.(2019)Ruder, Vulic, and S{\o}gaard]{Ruder2019}
Sebastian Ruder, Ivan Vulic, and Anders S{\o}gaard.
\newblock A survey of cross-lingual word embedding models.
\newblock \emph{J. Artif. Intell. Res.}, 65:\penalty0 569--631, 2019.
\newblock \doi{10.1613/jair.1.11640}.
\newblock URL \url{https://doi.org/10.1613/jair.1.11640}.

\bibitem[Ruder et~al.(2021)Ruder, Constant, Botha, Siddhant, Firat, Fu, Liu,
  Hu, Neubig, and Johnson]{xtreme-r}
Sebastian Ruder, Noah Constant, Jan Botha, Aditya Siddhant, Orhan Firat, Jinlan
  Fu, Pengfei Liu, Junjie Hu, Graham Neubig, and Melvin Johnson.
\newblock {XTREME-R:} towards more challenging and nuanced multilingual
  evaluation.
\newblock \emph{CoRR}, abs/2104.07412, 2021.
\newblock URL \url{https://arxiv.org/abs/2104.07412}.

\bibitem[Siddhant et~al.(2020)Siddhant, Johnson, Tsai, Ari, Riesa, Bapna,
  Firat, and Raman]{MMTE}
Aditya Siddhant, Melvin Johnson, Henry Tsai, Naveen Ari, Jason Riesa, Ankur
  Bapna, Orhan Firat, and Karthik Raman.
\newblock Evaluating the cross-lingual effectiveness of massively multilingual
  neural machine translation.
\newblock In \emph{The Thirty-Fourth {AAAI} Conference on Artificial
  Intelligence, {AAAI} 2020, The Thirty-Second Innovative Applications of
  Artificial Intelligence Conference, {IAAI} 2020, The Tenth {AAAI} Symposium
  on Educational Advances in Artificial Intelligence, {EAAI} 2020, New York,
  NY, USA, February 7-12, 2020}, pp.\  8854--8861. {AAAI} Press, 2020.
\newblock URL \url{https://aaai.org/ojs/index.php/AAAI/article/view/6414}.

\bibitem[Singh et~al.(2019)Singh, McCann, Keskar, Xiong, and
  Socher]{Jasdeep2019}
Jasdeep Singh, Bryan McCann, Nitish~Shirish Keskar, Caiming Xiong, and Richard
  Socher.
\newblock {XLDA:} cross-lingual data augmentation for natural language
  inference and question answering.
\newblock \emph{CoRR}, abs/1905.11471, 2019.
\newblock URL \url{http://arxiv.org/abs/1905.11471}.

\bibitem[Verma et~al.(2019)Verma, Lamb, Beckham, Najafi, Mitliagkas,
  Lopez{-}Paz, and Bengio]{Verma2019}
Vikas Verma, Alex Lamb, Christopher Beckham, Amir Najafi, Ioannis Mitliagkas,
  David Lopez{-}Paz, and Yoshua Bengio.
\newblock Manifold mixup: Better representations by interpolating hidden
  states.
\newblock In Kamalika Chaudhuri and Ruslan Salakhutdinov (eds.),
  \emph{Proceedings of the 36th International Conference on Machine Learning,
  {ICML} 2019, 9-15 June 2019, Long Beach, California, {USA}}, volume~97 of
  \emph{Proceedings of Machine Learning Research}, pp.\  6438--6447. {PMLR},
  2019.
\newblock URL \url{http://proceedings.mlr.press/v97/verma19a.html}.

\bibitem[Vincent et~al.(2008)Vincent, Larochelle, Bengio, and
  Manzagol]{vincent2008extracting}
Pascal Vincent, Hugo Larochelle, Yoshua Bengio, and Pierre-Antoine Manzagol.
\newblock Extracting and composing robust features with denoising autoencoders.
\newblock In \emph{Proceedings of the 25th international conference on Machine
  learning}, pp.\  1096--1103, 2008.

\bibitem[Wan et~al.(2011)Wan, Pan, and Li]{Wan2011}
C.~Wan, Rong Pan, and Jiefei Li.
\newblock Bi-weighting domain adaptation for cross-language text
  classification.
\newblock In \emph{IJCAI}, 2011.

\bibitem[Wei et~al.(2020)Wei, Hu, Weng, Xing, Yu, and Luo]{hictl}
Xiangpeng Wei, Yue Hu, Rongxiang Weng, Luxi Xing, Heng Yu, and Weihua Luo.
\newblock On learning universal representations across languages.
\newblock \emph{CoRR}, abs/2007.15960, 2020.
\newblock URL \url{https://arxiv.org/abs/2007.15960}.

\bibitem[Wu \& Dredze(2020)Wu and Dredze]{Wu2020}
Shijie Wu and Mark Dredze.
\newblock Are all languages created equal in multilingual {BERT}?
\newblock In \emph{Proceedings of the 5th Workshop on Representation Learning
  for NLP}, pp.\  120--130, Online, July 2020. Association for Computational
  Linguistics.
\newblock \doi{10.18653/v1/2020.repl4nlp-1.16}.
\newblock URL \url{https://www.aclweb.org/anthology/2020.repl4nlp-1.16}.

\bibitem[Xue et~al.(2020)Xue, Constant, Roberts, Kale, Al{-}Rfou, Siddhant,
  Barua, and Raffel]{mt5}
Linting Xue, Noah Constant, Adam Roberts, Mihir Kale, Rami Al{-}Rfou, Aditya
  Siddhant, Aditya Barua, and Colin Raffel.
\newblock mt5: {A} massively multilingual pre-trained text-to-text transformer.
\newblock \emph{CoRR}, abs/2010.11934, 2020.
\newblock URL \url{https://arxiv.org/abs/2010.11934}.

\bibitem[Yang et~al.(2020)Yang, Ma, Zhang, Wu, Li, and Zhou]{Yang2020}
Jian Yang, Shuming Ma, Dongdong Zhang, Shuangzhi Wu, Zhoujun Li, and Ming Zhou.
\newblock Alternating language modeling for cross-lingual pre-training.
\newblock In \emph{The Thirty-Fourth {AAAI} Conference on Artificial
  Intelligence, {AAAI} 2020, The Thirty-Second Innovative Applications of
  Artificial Intelligence Conference, {IAAI} 2020, The Tenth {AAAI} Symposium
  on Educational Advances in Artificial Intelligence, {EAAI} 2020, New York,
  NY, USA, February 7-12, 2020}, pp.\  9386--9393. {AAAI} Press, 2020.
\newblock URL \url{https://aaai.org/ojs/index.php/AAAI/article/view/6480}.

\bibitem[Yang et~al.(2019)Yang, Zhang, Tar, and Baldridge]{pawsx}
Yinfei Yang, Yuan Zhang, Chris Tar, and Jason Baldridge.
\newblock {PAWS}-{X}: A cross-lingual adversarial dataset for paraphrase
  identification.
\newblock In \emph{Proceedings of the 2019 Conference on Empirical Methods in
  Natural Language Processing and the 9th International Joint Conference on
  Natural Language Processing (EMNLP-IJCNLP)}, pp.\  3687--3692, Hong Kong,
  China, November 2019. Association for Computational Linguistics.
\newblock \doi{10.18653/v1/D19-1382}.
\newblock URL \url{https://www.aclweb.org/anthology/D19-1382}.

\bibitem[Zhang et~al.(2018)Zhang, Ciss{\'{e}}, Dauphin, and
  Lopez{-}Paz]{Zhang2018}
Hongyi Zhang, Moustapha Ciss{\'{e}}, Yann~N. Dauphin, and David Lopez{-}Paz.
\newblock mixup: Beyond empirical risk minimization.
\newblock In \emph{6th International Conference on Learning Representations,
  {ICLR} 2018, Vancouver, BC, Canada, April 30 - May 3, 2018, Conference Track
  Proceedings}. OpenReview.net, 2018.
\newblock URL \url{https://openreview.net/forum?id=r1Ddp1-Rb}.

\bibitem[Zhao et~al.(2020)Zhao, Eger, Bjerva, and Augenstein]{zhao2020}
Wei Zhao, Steffen Eger, Johannes Bjerva, and Isabelle Augenstein.
\newblock Inducing language-agnostic multilingual representations.
\newblock \emph{CoRR}, abs/2008.09112, 2020.
\newblock URL \url{https://arxiv.org/abs/2008.09112}.

\bibitem[Zheng et~al.(2021)Zheng, Dong, Huang, Wang, Chi, Singhal, Che, Liu,
  Song, and Wei]{zheng2021}
Bo~Zheng, Li~Dong, Shaohan Huang, Wenhui Wang, Zewen Chi, Saksham Singhal,
  Wanxiang Che, Ting Liu, Xia Song, and Furu Wei.
\newblock Consistency regularization for cross-lingual fine-tuning.
\newblock In \emph{Proceedings of the 59th Annual Meeting of the Association
  for Computational Linguistics and the 11th International Joint Conference on
  Natural Language Processing (Volume 1: Long Papers)}, pp.\  3403--3417,
  Online, August 2021. Association for Computational Linguistics.
\newblock \doi{10.18653/v1/2021.acl-long.264}.
\newblock URL \url{https://aclanthology.org/2021.acl-long.264}.

\bibitem[Zhou et~al.(2019)Zhou, Zhang, Jin, Zhu, Fang, Goh, and Kwok]{Zhou2019}
Joey~Tianyi Zhou, Hao Zhang, Di~Jin, Hongyuan Zhu, Meng Fang, Rick Siow~Mong
  Goh, and Kenneth Kwok.
\newblock Dual adversarial neural transfer for low-resource named entity
  recognition.
\newblock In \emph{Proceedings of the 57th Annual Meeting of the Association
  for Computational Linguistics}, pp.\  3461--3471, Florence, Italy, July 2019.
  Association for Computational Linguistics.
\newblock \doi{10.18653/v1/P19-1336}.
\newblock URL \url{https://www.aclweb.org/anthology/P19-1336}.

\end{thebibliography}
\bibliographystyle{iclr2022_conference}

\clearpage
\appendix

\section{Method Details}
\subsection{Multi-head Attention}
\label{multi_head}
In the multi-head attention layer, multiple attention heads are concatenated
\begin{equation}
\text{MultiHead}(Q,K,V)=\text{Concat}(\text{head}_{1, ..., h})\bm{W}^O,
\end{equation}
and each head is the scaled dot-product attention
\begin{equation}
\text{head}_{i}=\text{Attention}(Q\bm{W}_{i}^{q},K\bm{W}_{i}^{k},V\bm{W}_{i}^{v}),
\end{equation}
\begin{equation}
\text{Attention}(Q,K,V)=\text{softmax}(\frac{QK^\top}{\sqrt{d}})V,
\end{equation}
where $\bm{W}^O$, $\bm{W}^q$, $\bm{W}^k$ and $\bm{W}^v$ are trainable parameters.

\subsection{Training Objective}
\label{task_loss}
For classification tasks (e.g. NLI), the task loss is the cross-entropy loss
\begin{equation}
\mathcal{L}_{\text{task}}=-\sum_j^C \bm{y}_j \text{log} \bm{p}_j,
\end{equation}
where $C$ is the size of the label set.

For structured prediction tasks (e.g. POS) and span extraction tasks (e.g. QA), the task loss is also the cross-entropy loss
\begin{equation}
\mathcal{L}_{\text{task}}=-\sum_i^n \sum_j^C \bm{y}_{ij} \text{log} \bm{p}_{ij}.
\end{equation}
where $n$ is the sequence length. 

\clearpage
\section{Analysis Results}
\label{appendix_analysis}
\begin{figure}[h]
\centering
\subfigure[PAWS-X]{
\label{fig_correl_a}
\includegraphics[width=0.35\textwidth]{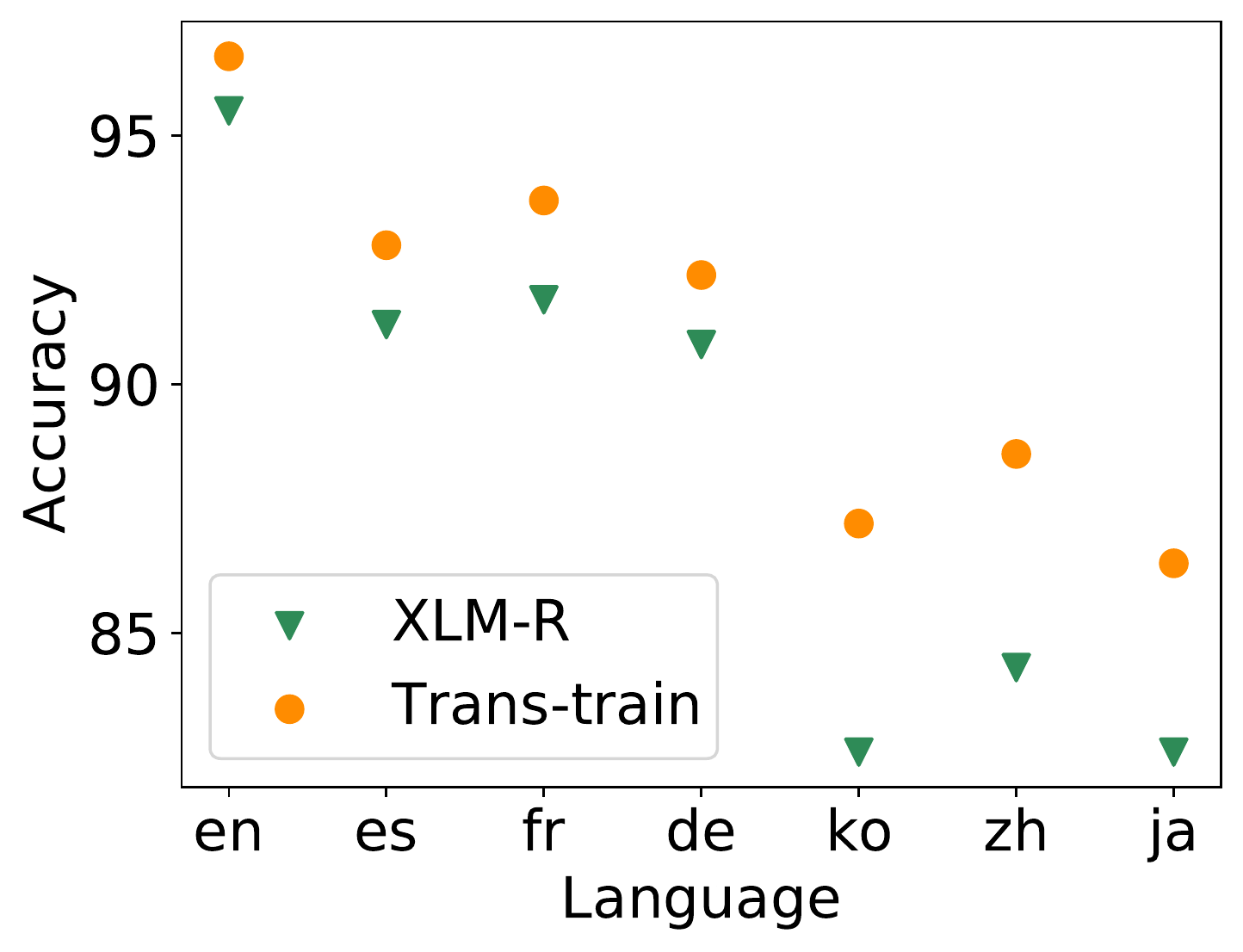}}
\qquad
\subfigure[XNLI]{
\label{fig_correl_b} 
\includegraphics[width=0.35\textwidth]{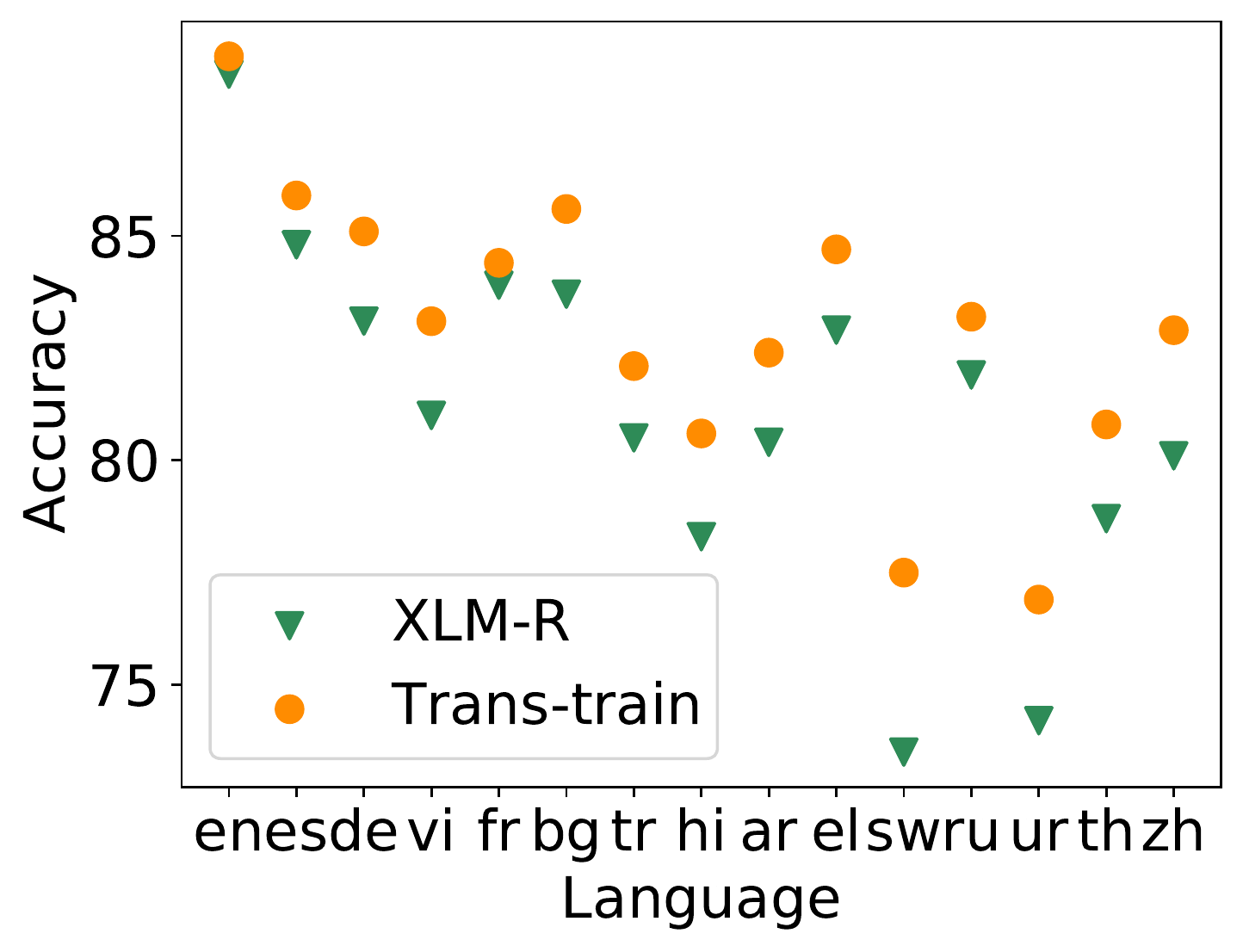}}
\caption{Performances on PAWS-X and XNLI test set, where languages are sorted by decreasing CKA scores. The trend indicates the performance gets worse along with the CKA score decreasing.
}
\label{fig_cka_line}
\end{figure}
\begin{figure}[ht]
\centering
\subfigure[w/o \method]{
\label{fig_cka_a}
\includegraphics[width=0.35\textwidth]{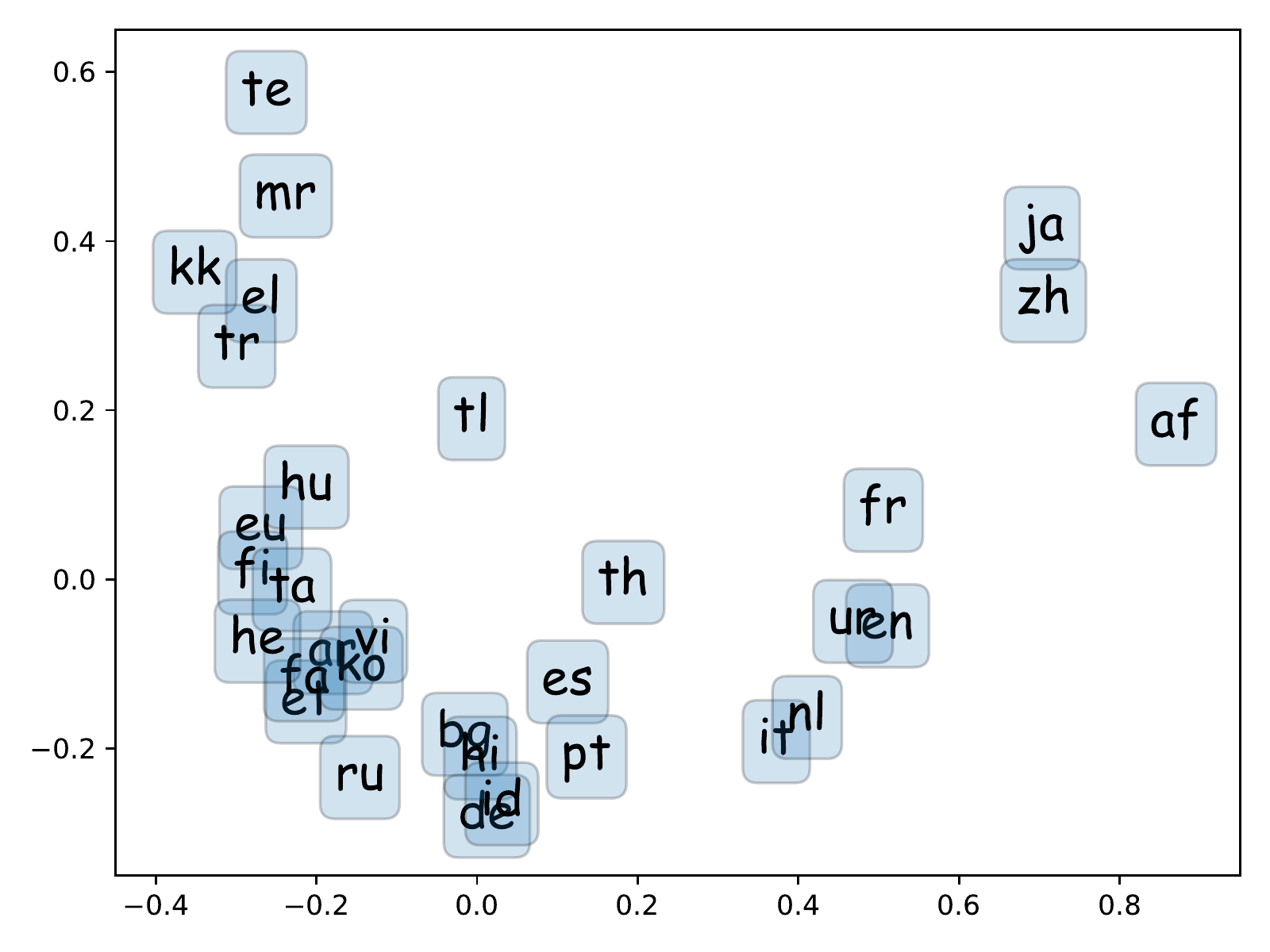}}
\qquad
\subfigure[\method]{
\label{fig_cka_b} 
\includegraphics[width=0.35\textwidth]{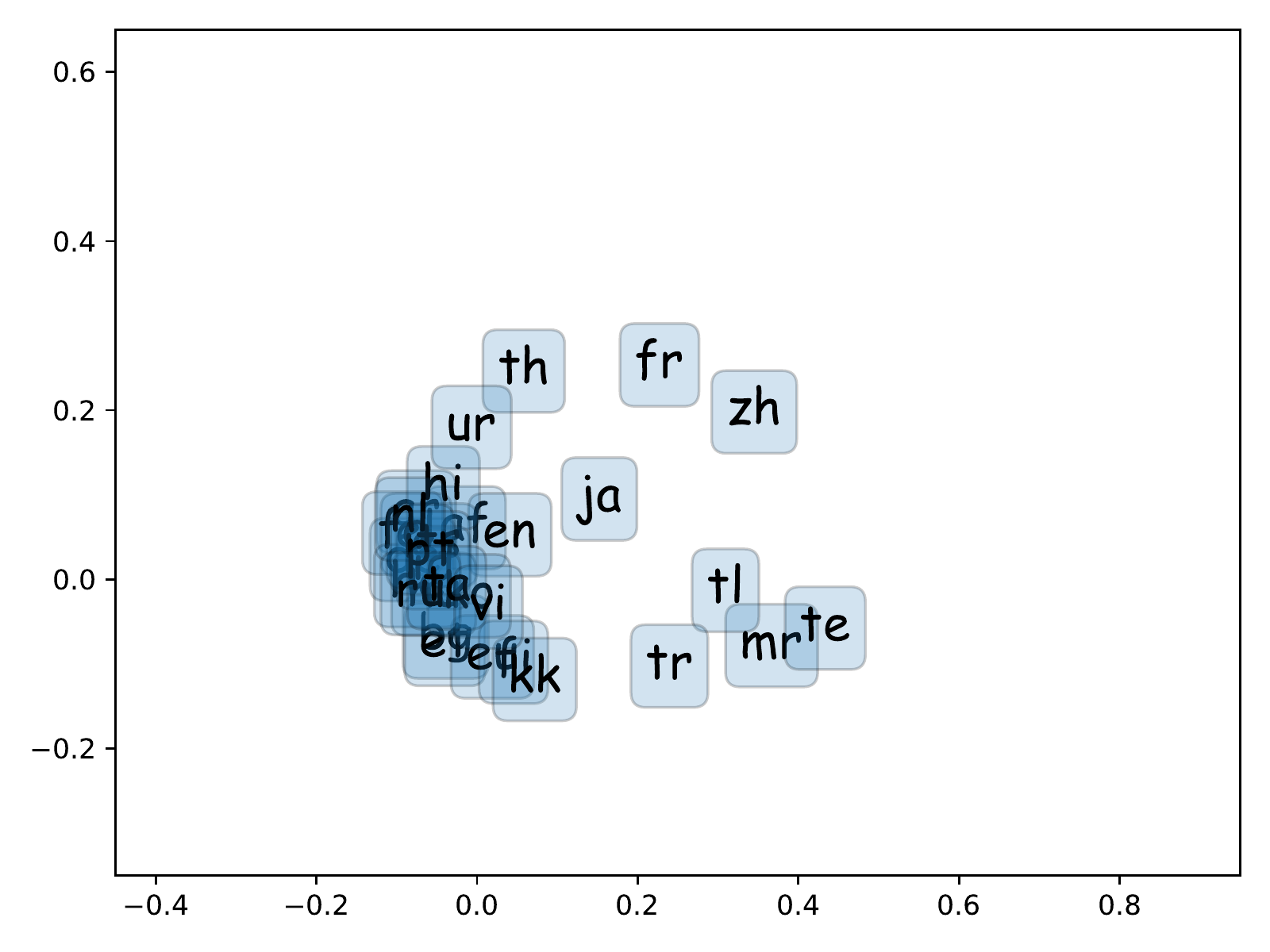}}
\caption{Language centroids visualization of the POS test set, which indicates X-Mixup brings closer these centroids obviously.  
} %
\label{fig_cka_rep}
\end{figure}
\begin{figure}[ht]
\centering
\includegraphics[scale=0.35]{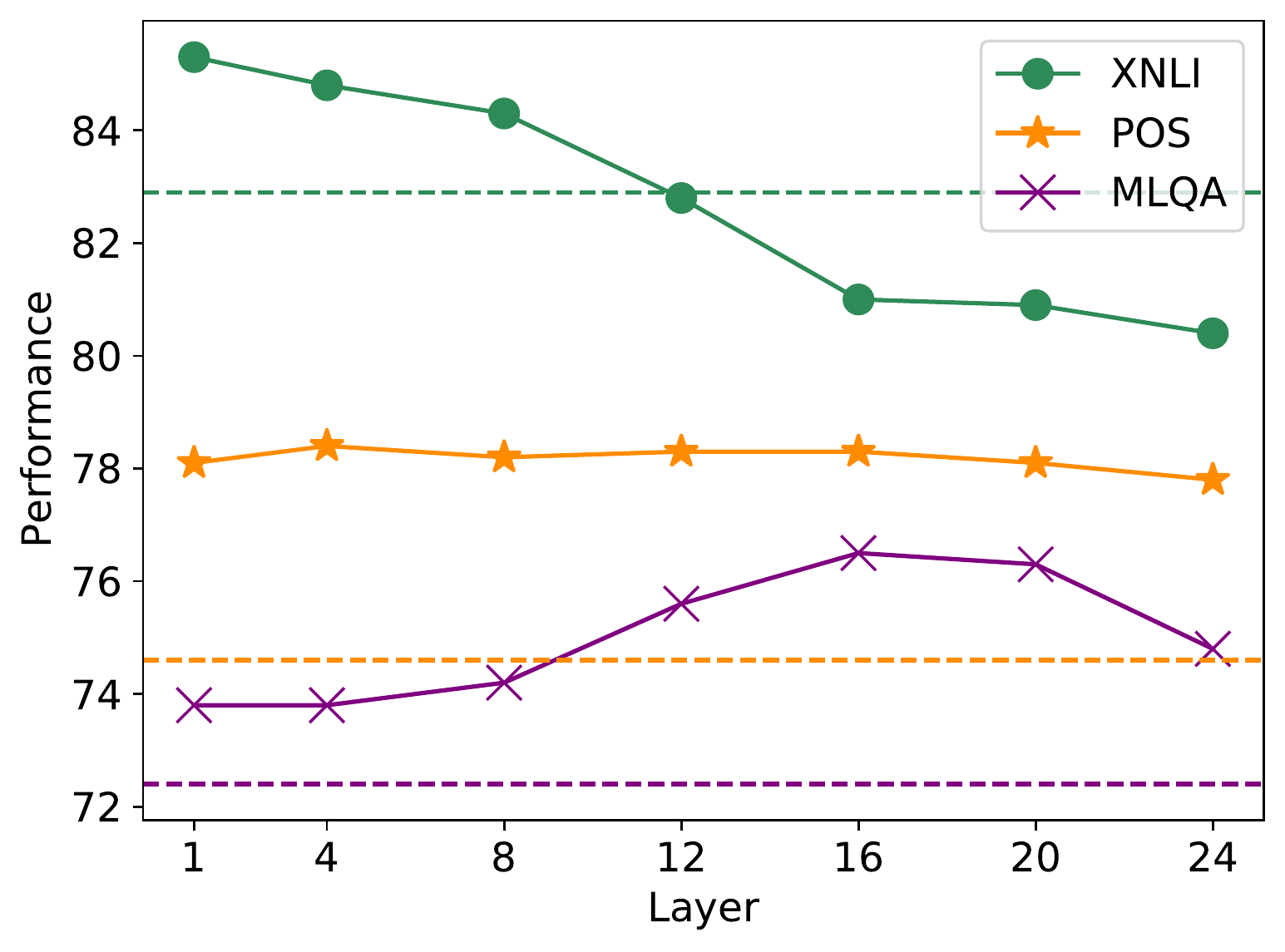}
\caption{Performances on implementing \method (solid line) in different layers and Trans-train (dashed line) on three downstream tasks.
}
\label{fig_layer}
\end{figure}
\begin{table*}[ht]
\setlength{\belowcaptionskip}{0.2cm}
\caption{Performances on XNLI test set, where Trans-train and \method are trained on 8 seen languages and tested on both these seen languages and 7 unseen languages. $\Delta$ is the performance difference between \method and Trans-train. 
Results show \method performs better than Trans-train by a large margin on both seen and unseen languages.
}
\centering
\resizebox{\linewidth}{!}{
\begin{tabular}{l|cccccccc|ccccccc|c}
\toprule
\multirow{2}{*}{Model} & \multicolumn{8}{c|}{Seen Languages} & \multicolumn{7}{c|}{Unseen Languages} & \multirow{2}{*}{Avg.} \\
 & en & bg & el & fr & ru & th & ur & zh & ar & de & es & hi & sw & tr & vi &  \\
\hline
Trans-train & 87.6 & 84.7 & 84.2 & 84.6 & 82.9 & 80.1 & 76.5 & 83.0 & 82.6 & 84.5 & 85.0 & 80.6 & 77.8 & 82.7 & 82.7 & 82.6 \\
\method & 89.5 & 87.1 & 86.3 & 86.8 & 84.7 & 82.7 & 79.0 & 85.0 & 85.3 &  86.3 & 86.9 & 82.9 & 80.3 & 84.5 & 84.5 & 84.8 \\
$\Delta$ & +1.9 & +2.4 & +1.9 & +2.2 & +1.8 & +2.6 & +2.5 & +2.0 & +2.7 & +1.8 & +1.9 & +1.7 & +2.5 & +1.8 & +1.8 & +1.8 \\
\bottomrule
\end{tabular}
}
\label{table_xnli_new_lang}
\end{table*}
\begin{table*}[ht]
\setlength{\belowcaptionskip}{0.2cm}
\caption{CKA scores and performances on XNLI test set, where $\Delta$ is the score or performance difference between \method and Trans-train.
Results show \method improves the CKA scores evenly across different target languages and the performance improvements are diverse. There is no obvious correlation between the CKA score improvement and performance improvement.
}
\centering
\resizebox{\linewidth}{!}{
\begin{tabular}{lcccccccccccccccc}
\toprule
Model & en & es & de & vi & fr & bg & tr & el & ru & ar & hi & sw & ur & th & zh & Avg. \\
\midrule
\textsl{CKA score} & & & & & & & & \\
Trans-train & 1.00 & 0.81 & 0.80 & 0.78 & 0.78 & 0.78 & 0.76 & 0.75 & 0.75 & 0.74 & 0.74 & 0.73 & 0.72 & 0.72 & 0.72 & 0.77 \\
\method & 1.00 & 0.88 & 0.87 & 0.86 & 0.86 & 0.86 & 0.85 & 0.83 & 0.82 & 0.84 & 0.83 & 0.82 & 0.81 & 0.80 & 0.79 & 0.85 \\
$\Delta$ & 0.00 & 0.07 & 0.07 & 0.08 & 0.08 & 0.08 & 0.09 & 0.08 & 0.07 & 0.10 & 0.09 & 0.09 & 0.09 & 0.08 & 0.07 & 0.08 \\
\midrule
\textsl{Performance} & & & & & & & & \\
Trans-train & 88.6 & 85.7 & 84.5 & 82.6 & 84.2 & 85.2 & 82.1 & 84.5 & 81.8 & 82.2 & 80.8 & 77.0 & 77.7 & 80.2 & 82.7 & 82.6 \\
\method & 89.9 & 87.7 & 86.9 & 85.4 & 87.1 & 87.3 & 84.9 & 86.8 & 85.1 &  85.2 & 83.5 & 81.2 & 79.6 & 83.2 & 85.2 & 85.3 \\
$\Delta$ & +1.3 & +2.0 & +2.4 & +2.8 & +2.9 & +2.1 & +2.8 & +2.3 & +3.3 & +3.0 & +2.7 & +4.2 & +1.9 & +3.0 & +2.5 & +2.7 \\
\bottomrule
\end{tabular}
}
\label{table_xnli_cka}
\end{table*}
\begin{table*}[ht]
\setlength{\belowcaptionskip}{0.2cm}
\caption{The cross-lingual transfer gap (lower is better) of different methods on the XTREME benchmark. For QA tasks, we only show EM scores. $^{\dagger}$ denotes results from \citet{hictl}. 
Overall, \method achieves the smallest cross-lingual transfer gap on four out of seven datasets.
}
\centering
\resizebox{\linewidth}{!}{
\begin{tabular}{l c c c c c c c c}
\toprule
Model & XNLI & PAWS-X & POS & NER & XQuAD & MLQA & TyDiQA & Avg. \\
\midrule
mBERT \citep{xtreme} & 16.5 & 14.1 & 25.5 & 23.6 & 25.0 & 27.5 & 22.2 & 22.1 \\
XLM-R \citep{xtreme} & 10.2 & 12.4 & 24.3 & 19.8 & 16.3 & 19.1 & 13.1 & 16.5 \\
Trans-train \citep{xtreme} & 7.3 & 9.0 & 22.4$^{\dagger}$ & 20.5$^{\dagger}$ & 17.6 & 22.2 & 24.2 & 17.6 \\
Filter \citep{Filter} & 6.0 & \textbf{5.2} & 19.7 & 16.3 & 7.3 & 15.7 & 9.2 & 11.3 \\
\textsc{xTune} \citep{zheng2021} & 5.5 & \textbf{5.2} & \textbf{17.3} & \textbf{14.8} & 10.1 & 18.5 & \textbf{0.9} & \textbf{10.3} \\
\midrule
\method & \textbf{4.9} & \textbf{5.2} & 18.1 & 15.9 & \textbf{6.7} & \textbf{13.9} & 9.6 & 10.6 \\
\bottomrule
\end{tabular}
}
\label{table_gap}
\end{table*}
\begin{table}[t]
\footnotesize
\setlength{\belowcaptionskip}{0.2cm}
\centering
\caption{Ablation results on the consistency loss, which show the KL consistency loss contributes more than the MSE consistency loss on the classification task.}
\begin{tabular}{l c c c}
\toprule
Model & XNLI & POS & MLQA \\
\midrule
\method & 85.3 & 78.4 & 76.5/58.1 \\
\quad w/o MSE consistency loss & 84.6 & 78.0 & 76.5/58.0 \\
\quad w/o KL consistency loss & 84.3 & - & - \\
\quad w/o both & 84.2 & - & - \\
\bottomrule
\end{tabular}
\label{table_ablation_consist}
\end{table}

\clearpage
\section{Experimental Details}
\subsection{Hyper-parameters}
\label{hyper}
For all tasks, we fine-tune on 8 Nvidia V100-32GB GPU cards with the batch size 64.
For XQuAD and MLQA, we finetune 2 epochs. For other tasks, we finetune 4 epochs. 
There is no dev set in XQuAD, so we use the dev set of MLQA for the model selection.
Table \ref{table_hyper} shows hyper-parameters used for \method.
\begin{table}[ht]
\footnotesize
\setlength{\belowcaptionskip}{0.2cm}
\centering
\caption{Hyper-parameters used for \method, where $\alpha$ is used for balanced training in Eq \ref{final_task_loss} and $p_k$ is the scheduled sampling decay rate.}
\begin{tabular}{l c c c}
\toprule
Parameter & Classification & Structured Prediction & QA \\
\midrule
$\alpha$ & 0.4 & 0.8 & 0.2 \\
$p_k$ & 1000 & 1000 & 2000 \\
\bottomrule
\end{tabular}
\label{table_hyper}
\end{table}

\subsection{Detailed Results}
\label{detail_result}
Detailed results of each tasks and languages are shown below.
Results of mBERT, XLM, MMTE and XLM-R are from XTREME \citep{xtreme}. Results of Filter is the best results of \citet{Filter}.
\begin{table*}[ht]
\centering
\resizebox{\linewidth}{!}{
\begin{tabular}{lcccccccccccccccc}
\toprule
{\bf Model} & {\bf en} & {\bf ar} & {\bf bg} & {\bf de} & {\bf el} & {\bf es} & {\bf fr} & {\bf hi} & {\bf ru} & {\bf sw} & {\bf th} & {\bf tr} & {\bf ur} & {\bf vi} & {\bf zh} & {\bf Avg.} \\
\midrule
mBERT & 81.9 & 73.8 & 77.6 & 77.6 & 75.9 & 79.1 & 77.8 & 70.7 & 75.4 & 70.5 & 70.0 & 74.3 & 67.4 & 77.0 & 77.6 & 75.1 \\
XLM         & 82.8          & 66.0          & 71.9          & 72.7          & 70.4          & 75.5          & 74.3          & 62.5          & 69.9          & 58.1          & 65.5          & 66.4          & 59.8          & 70.7          & 70.2          & 69.1          \\
MMTE        & 79.6          & 64.9          & 70.4          & 68.2          & 67.3          & 71.6          & 69.5          & 63.5          & 66.2          & 61.9          & 66.2          & 63.6          & 60.0          & 69.7          & 69.2          & 67.5         \\
XLM-R & 88.6 & 82.2 & 85.2 & 84.5 & 84.5 & 85.7 & 84.2 & 80.8 & 81.8 & 77.0 & 80.2 & 82.1 & 77.7 & 82.6 & 82.7 & 82.6 \\
Filter & 89.5 & 83.6 & 86.4 & 85.6 & 85.4 & 86.6 & 85.7 & 81.1 & 83.7 & 78.7 & 81.7 & 83.2 & 79.1 & 83.9 & 83.8 & 83.9 \\
\textsc{xTune} & 89.9 & 84.0 & 87.0 & 86.5 & 86.2 & 87.4 & 86.6 & 83.2 & 85.2 & 80.0 & 82.7 & 84.1 & 79.6 & 84.8 & 84.3 & 84.8 \\
\midrule
\method & 89.9 & 85.2 & 87.3 & 86.9 & 86.8 & 87.7 & 87.1 & 83.5 & 85.1 & 81.2 & 83.2 & 84.9 & 79.6 & 85.4 & 85.2 & 85.3 \\
\bottomrule
\end{tabular}
}
\caption{XNLI accuracy scores for each language.}
\label{table_xnli}
\end{table*}
\begin{table*}[ht]
\centering
\begin{tabular}{lcccccccc}
\toprule
{\bf Model} & {\bf en} & {\bf de} & {\bf es} & {\bf fr} & {\bf ja} & {\bf ko} & {\bf zh} & {\bf Avg.} \\
\midrule
mBERT & 94.0 & 85.7 & 87.4 & 87.0 & 73.0 & 69.6 & 77.0 & 81.9\\
XLM & 94.0& 85.9& 88.3& 87.4& 69.3& 64.8& 76.5 & 80.9 \\
MMTE & 93.1& 85.1& 87.2& 86.9& 72.0& 69.2& 75.9 & 81.3 \\
XLM-R & 94.7& 89.7& 90.1& 90.4& 78.7& 79.0& 82.3 & 86.4 \\
Filter & 95.9 & 92.8 & 93.0 & 93.7 & 87.4 & 87.6 & 89.6 & 91.5 \\
\textsc{xTune} & 96.1 & 92.6 & 93.1 & 93.9 & 87.8 & 89.0 & 88.8 & 91.6 \\
\midrule
\method & 96.3 & 93.2 & 93.6 & 94.6 & 87.3 & 88.2 & 89.5 & 91.8 \\
\bottomrule
\end{tabular}
\caption{PAWS-X accuracy scores for each language.}
\label{table_pawsx}
\end{table*}

\begin{table*}[ht]
\centering
\resizebox{\linewidth}{!}{
\begin{tabular}{lccccccccccccccccc}
\toprule
{\bf Model} & {\bf af} & {\bf ar} & {\bf bg} & {\bf de} & {\bf el} & {\bf en} & {\bf es} & {\bf et} & {\bf eu} & {\bf fa} & {\bf fi} & {\bf fr} & {\bf he} & {\bf hi} & {\bf hu} & {\bf id} & {\bf it} \\
\midrule
mBERT & 86.6& 56.2& 85.0& 85.2& 81.1& 95.5& 86.9& 79.1& 60.7& 66.7& 78.9& 84.2& 56.2& 67.2& 78.3& 71.0& 88.4 \\
XLM   & 88.5          & 63.1          & 85.0          & 85.8          & 84.3          & 95.4          & 85.8          & 78.3          & 62.8          & 64.7          & 78.4          & 82.8          & 65.9          & 66.2          & 77.3          & 70.2          & 87.4          \\
XLM-R & 89.8& 67.5& 88.1& 88.5& 86.3& 96.1& 88.3& 86.5& 72.5& 70.6& 85.8& 87.2& 68.3& 76.4& 82.6& 72.4& 89.4 \\
Filter & 88.7 & 66.1 & 88.5 & 89.2 & 88.3 & 96.0 & 89.1 & 86.3 & 78.0 & 70.8 & 86.1 & 88.9 & 64.9 & 76.7 & 82.6 & 72.6 & 89.8 \\
\textsc{xTune} & 90.4 & 72.8 & 89.0 & 89.4 & 87.0 & 96.1 & 88.8 & 88.1 & 73.1 & 74.7 & 87.2 & 89.5 & 83.5 & 77.7 & 83.6 & 73.2 & 90.5 \\
\midrule
\method & 89.4 & 70.1 & 88.8 & 88.7 & 86.7 & 96.0 & 89.0 & 88.3 & 76.2 & 72.5 & 87.0 & 88.2 & 82.4 & 78.0 & 83.8 & 72.4 & 90.3 \\
\midrule
{\bf Model} & {\bf ja} & {\bf kk} & {\bf ko} & {\bf mr} & {\bf nl} & {\bf pt} & {\bf ru} & {\bf ta} & {\bf te} & {\bf th} & {\bf tl} & {\bf tr} & {\bf ur} & {\bf vi} & {\bf yo} & {\bf zh} & {\bf Avg.} \\
\midrule
mBERT & 49.2 & 70.5 & 49.6 & 69.4 & 88.6 & 86.2 & 85.5 & 59.0 & 75.9 & 41.7 & 81.4 & 68.5 & 57.0 & 53.2 & 55.7 & 61.6 & 71.5 \\
XLM   & 49.0          & 70.2          & 50.1          & 68.7          & 88.1          & 84.9          & 86.5          & 59.8          & 76.8          & 55.2          & 76.3          & 66.4          & 61.2          & 52.4          & 20.5          & 65.4          & 71.3          \\
XLM-R & 15.9 & 78.1 & 53.9 & 80.8 & 89.5 & 87.6 & 89.5 & 65.2 & 86.6 & 47.2 & 92.2 & 76.3 & 70.3 & 56.8 & 24.6 & 25.7 & 73.8 \\
Filter & 40.4 & 80.4 & 53.3 & 86.4 & 89.4 & 88.3 & 90.5 & 65.3 & 87.3 & 57.2 & 94.1 & 77.0 & 70.9 & 58.0 & 43.1 & 53.1 & 76.9 \\
\textsc{xTune} & 65.3 & 79.8 & 56.0 & 85.5 & 89.7 & 89.3 & 90.8 & 65.7 & 85.5 & 61.4 & 93.8 & 78.3 & 74.0 & 57.5 & 27.9 & 68.8 & 79.3 \\
\midrule
\method & 62.7 & 79.0 & 55.3 & 84.8 & 89.6 & 88.8 & 90.1 & 63.6 & 87.4 & 59.9 & 93.1 & 77.1 & 72.4 & 59.4 & 27.3 & 68.3 & 78.4 \\
\bottomrule
\end{tabular}
}
\caption{POS results (F1) for each language.}
\label{table_pos}
\end{table*}
\begin{table*}[ht]
\centering
\resizebox{\linewidth}{!}{
\begin{tabular}{lcccccccccccccccccccc}
\toprule
{\bf Model} & {\bf en} & {\bf af} & {\bf ar} & {\bf bg} & {\bf bn} & {\bf de} & {\bf el} & {\bf es} & {\bf et} & {\bf eu} & {\bf fa} & {\bf fi} & {\bf fr} & {\bf he} & {\bf hi} & {\bf hu} & {\bf id} & {\bf it} & {\bf ja} & {\bf jv} \\
\midrule
mBERT & 85.2 & 77.4 & 41.1 & 77.0 & 70.0 & 78.0 & 72.5 & 77.4 & 75.4 & 66.3 & 46.2 & 77.2 & 79.6 & 56.6 & 65.0 & 76.4 & 53.5 & 81.5 & 29.0 & 66.4 \\
XLM & 82.6 & 74.9 & 44.8 & 76.7 & 70.0 & 78.1 & 73.5 & 74.8 & 74.8 & 62.3 & 49.2 & 79.6 & 78.5 & 57.7 & 66.1 & 76.5 & 53.1 & 80.7 & 23.6 & 63.0 \\
MMTE & 77.9 & 74.9 & 41.8 & 75.1 & 64.9 & 71.9 & 68.3 & 71.8 & 74.9 & 62.6 & 45.6 & 75.2 & 73.9 & 54.2 & 66.2 & 73.8 & 47.9 & 74.1 & 31.2 & 63.9 \\
XLM-R & 84.7 & 78.9 & 53.0 & 81.4 & 78.8 & 78.8 & 79.5 & 79.6 & 79.1 & 60.9 & 61.9 & 79.2 & 80.5 & 56.8 & 73.0 & 79.8 & 53.0 & 81.3 & 23.2 & 62.5 \\
Filter & 83.5 & 80.4 & 60.7 & 83.5 & 78.4 & 80.4 & 80.7 & 74.0 & 81.0 & 66.9 & 71.3 & 80.2 & 79.9 & 57.4 & 74.3 & 82.2 & 54.0 & 81.9 & 24.3 & 63.5 \\
\textsc{xTune} & 85.0 & 80.4 & 59.1 & 84.8 & 79.1 & 80.5 & 82.0 & 78.1 & 81.5 & 64.5 & 65.9 & 82.2 & 81.9 & 62.0 & 75.0 & 82.8 & 55.8 & 83.1 & 30.5 & 65.9 \\
\midrule
\method & 84.5 & 79.0 & 58.4 & 84.0 & 81.4 & 80.6 & 81.4 & 73.8 & 81.5 & 65.7 & 61.6 & 80.4 & 80.3 & 64.4 & 74.7 & 82.0 & 53.4 & 82.2 & 38.8 & 63.5 \\
\midrule
{\bf Model}  & {\bf ka} & {\bf kk} & {\bf ko} & {\bf ml} & {\bf mr} & {\bf ms} & {\bf my} & {\bf nl} & {\bf pt} & {\bf ru} & {\bf sw} & {\bf ta} & {\bf te} & {\bf th} & {\bf tl} & {\bf tr} & {\bf ur} & {\bf vi} & {\bf yo} & {\bf zh} \\
\midrule
mBERT & 64.6 & 45.8 & 59.6 & 52.3 & 58.2 & 72.7 & 45.2 & 81.8 & 80.8 & 64.0 & 67.5 & 50.7 & 48.5 & 3.6 & 71.7 & 71.8 & 36.9 & 71.8 & 44.9 & 42.7 \\
XLM & 67.7 & 57.2 & 26.3 & 59.4 & 62.4 & 69.6 & 47.6 & 81.2 & 77.9 & 63.5 & 68.4 & 53.6 & 49.6 & 0.3 & 78.6 & 71.0 & 43.0 & 70.1 & 26.5 & 32.4 \\
MMTE & 60.9 & 43.9 & 58.2 & 44.8 & 58.5 & 68.3 & 42.9 & 74.8 & 72.9 & 58.2 & 66.3 & 48.1 & 46.9 & 3.9 & 64.1 & 61.9 & 37.2 & 68.1 & 32.1 & 28.9 \\
XLMR & 71.6 & 56.2 & 60.0 & 67.8 & 68.1 & 57.1 & 54.3 & 84.0 & 81.9 & 69.1 & 70.5 & 59.5 & 55.8 & 1.3 & 73.2 & 76.1 & 56.4 & 79.4 & 33.6 & 33.1 \\
Filter & 71.0 & 51.1 & 63.8 & 70.2 & 69.8 & 69.3 & 59.0 & 84.6 & 82.1 & 71.1 & 70.6 & 64.3 & 58.7 & 2.4 & 74.4 & 83.0 & 73.4 & 75.8 & 42.9 & 35.4 \\
\textsc{xTune} & 76.3 & 56.9 & 67.1 & 72.6 & 71.5 & 72.5 & 66.7 & 85.8 & 82.1 & 75.2 & 72.4 & 66.0 & 61.8 & 1.1 & 77.5 & 83.7 & 75.6 & 80.8 & 44.9 & 36.5 \\
\midrule
\method & 76.5 & 51.7 & 63.9 & 69.8 & 71.2 & 70.4 & 67.9 & 84.5 & 83.1 & 73.5 & 70.7 & 65.6 & 59.3 & 4.4 & 75.0 & 81.8 & 73.1 & 78.2 & 41.6 & 47.8 \\
\bottomrule
\end{tabular}
}
\caption{NER results (F1) for each language.}
\label{table_ner}
\end{table*}
\begin{table*}[ht]
\centering
\resizebox{\linewidth}{!}{
\begin{tabular}{lcccccccccccc}
\toprule
{\bf Model} & {\bf en} & {\bf ar} & {\bf de} & {\bf el} & {\bf es} & {\bf hi} & {\bf ru} & {\bf th} & {\bf tr} & {\bf vi} & {\bf zh} & {\bf Avg.} \\
\midrule
mBERT & 83.5 / 72.2 &  61.5 / 45.1 &  70.6 / 54.0 &  62.6 / 44.9 &  75.5 / 56.9 &  59.2 / 46.0 &  71.3 / 53.3 &  42.7 / 33.5 &  55.4 / 40.1 &  69.5 / 49.6 &  58.0 / 48.3 &  64.5 / 49.4 \\
XLM & 74.2 / 62.1 &  61.4 / 44.7 &  66.0 / 49.7 &  57.5 / 39.1 &  68.2 / 49.8 &  56.6 / 40.3 &  65.3 / 48.2 &  35.4 / 24.5 &  57.9 / 41.2 &  65.8 / 47.6 &  49.7 / 39.7 &  59.8 / 44.3 \\
MMTE & 80.1 / 68.1 &  63.2 / 46.2 &  68.8 / 50.3 &  61.3 / 35.9 &  72.4 / 52.5 &  61.3 / 47.2 &  68.4 / 45.2 &  48.4 / 35.9 &  58.1 / 40.9 &  70.9 / 50.1 &  55.8 / 36.4 &  64.4 / 46.2 \\
XLM-R & 86.5 / 75.7 &  68.6 / 49.0 &  80.4 / 63.4 &  79.8 / 61.7 &  82.0 / 63.9 &  76.7 / 59.7 &  80.1 / 64.3 &  74.2 / 62.8 &  75.9 / 59.3 &  79.1 / 59.0 &  59.3 / 50.0 &  76.6 / 60.8 \\
Filter & 86.4 / 74.6 & 79.5 / 60.7 & 83.2 / 67.0 & 83.0 / 64.6 & 85.0 / 67.9 & 83.1 / 66.6 & 82.8 / 67.4 & 79.6 / 73.2 & 80.4 / 64.4 & 83.8 / 64.7 & 79.9 / 77.0 & 82.4 / 68.0 \\
\textsc{xTune} & 88.8 / 78.1 & 79.7 / 63.9 & 83.7 / 68.2 & 83.0 / 65.7  & 84.7 / 68.3 & 80.7 / 64.9 & 82.2 / 66.6 & 81.9 / 76.1 & 79.3 / 65.0 & 82.7 / 64.5 & 81.3 / 78.0 & 82.5 / 69.0 \\
\midrule
\method & 86.7 / 75.4 & 81.3 / 63.5 & 83.5 / 66.8 & 84.3 / 67.6 & 85.2 / 68.2 & 83.9 / 68.5 & 83.0 / 67.7 & 82.6 / 76.9 & 80.9 / 65.3 & 84.8 / 66.8 & 72.4 / 75.6 & 82.6 / 69.3 \\
\bottomrule
\end{tabular}
}
\caption{XQuAD results (F1 / EM) for each language.}
\label{table_xquad}
\end{table*}
\begin{table*}[ht]
\centering
\resizebox{\linewidth}{!}{
\begin{tabular}{lcccccccc}
\toprule
{\bf Model} & {\bf en} & {\bf ar} & {\bf de} & {\bf es} & {\bf hi} & {\bf vi} & {\bf zh} & {\bf Avg.} \\
\midrule
mBERT & 80.2 / 67.0 &  52.3 / 34.6 &  59.0 / 43.8 &  67.4 / 49.2 &  50.2 / 35.3 &  61.2 / 40.7 &  59.6 / 38.6 & 61.4 / 44.2\\
XLM & 68.6 / 55.2 &  42.5 / 25.2 &  50.8 / 37.2 &  54.7 / 37.9 &  34.4 / 21.1 &  48.3 / 30.2 &  40.5 / 21.9 & 48.5 / 32.6\\
MMTE & 78.5 / – &  56.1 / – &  58.4 / – &  64.9 / – &  46.2 / – &  59.4 / – &  58.3 / –  & 60.3 / 41.4\\
XLM-R & 83.5 / 70.6 &  66.6 / 47.1 &  70.1 / 54.9 &  74.1 / 56.6 &  70.6 / 53.1 &  74.0 / 52.9 &  62.1 / 37.0 & 71.6 / 53.2\\
Filter & 84.0 / 70.8 & 72.1 / 51.1 & 74.8 /60.0 & 78.1 / 60.1 & 76.0 / 57.6 & 78.1 /57.5 & 70.5 / 47.0 & 76.2 / 57.7\\
\textsc{xTune} & 85.3 / 72.9 & 69.7 / 50.1 & 72.3 / 57.3 & 76.3 / 58.8 & 74.0 / 56.0 & 76.5 / 55.9 & 70.8 / 48.3 & 75.0 / 57.1 \\
\midrule
\method & 83.1 / 70.0 & 71.9 / 51.1 & 74.5 / 59.4 & 77.7 / 60.0 & 76.3 / 57.7 & 78.0 / 57.5 & 73.7 / 51.1 & 76.5 / 58.1 \\
\bottomrule
\end{tabular}
}
\caption{MLQA results (F1 / EM) for each language.}
\label{table_mlqa}
\end{table*}
\begin{table*}[ht]
\centering
\resizebox{\linewidth}{!}{
\begin{tabular}{lcccccccccc}
\toprule
{\bf Model} & {\bf en} & {\bf ar} & {\bf bn} & {\bf fi} & {\bf id} & {\bf ko} & {\bf ru} & {\bf sw} & {\bf te} & {\bf Avg.} \\
\midrule
mBERT & 75.3 / 63.6 &  62.2 / 42.8 &  49.3 / 32.7 &  59.7 / 45.3 &  64.8 / 45.8 &  58.8 / 50.0 &  60.0 / 38.8 &  57.5 / 37.9 &  49.6 / 38.4 &  59.7 / 43.9 \\
XLM & 66.9 / 53.9 &  59.4 / 41.2 &  27.2 / 15.0 &  58.2 / 41.4 &  62.5 / 45.8 &  14.2 / 5.1 &  49.2 / 30.7 &  39.4 / 21.6 &  15.5 / 6.9 &  43.6 / 29.1 \\
MMTE & 62.9 / 49.8 &  63.1 / 39.2 &  55.8 / 41.9 &  53.9 / 42.1 &  60.9 / 47.6 &  49.9 / 42.6 &  58.9 / 37.9 &  63.1 / 47.2 &  54.2 / 45.8 &  58.1 / 43.8 \\
XLM-R & 71.5 / 56.8 &  67.6 / 40.4 &  64.0 / 47.8 &  70.5 / 53.2 &  77.4 / 61.9 &  31.9 / 10.9 &  67.0 / 42.1 &  66.1 / 48.1 &  70.1 / 43.6 &  65.1 / 45.0 \\
Filter & 72.4 / 59.1 & 72.8 / 50.8 & 70.5 / 56.6 & 73.3 / 57.2 & 76.8 / 59.8 & 33.1 / 12.3 & 68.9 / 46.6 & 77.4 / 65.7 & 69.9 / 50.4 & 68.3 / 50.9 \\
\textsc{xTune} & 73.8 / 61.6 & 77.8 / 60.2 & 73.5 / 61.1 & 77.0 / 62.2 & 80.8 / 68.1 & 66.9 / 56.5 & 72.1 / 51.9 & 77.9 / 65.3 & 77.6 / 60.7 & 75.3 / 60.8 \\
\midrule
\method & 73.9 / 61.4 & 73.8 / 54.2 & 67.4 / 49.6 & 75.4 / 60.6 & 78.8 / 65.0 & 32.9 / 12.0 & 69.1 / 52.2 & 78.0 / 66.9 & 72.0 / 53.5 & 69.0 / 52.8 \\
\bottomrule
\end{tabular}
}
\caption{TyDiQA-GolP results (F1 / EM) for each language.}
\label{table_tydiqa}
\end{table*}

\end{document}